\documentclass{article}

\usepackage{microtype}
\usepackage{graphicx}
\usepackage{subfigure}
\usepackage{booktabs} 
\usepackage{amsfonts}
\usepackage{amsthm}
\usepackage{amssymb}
\usepackage{amsmath}
\usepackage{hyperref}
\usepackage{graphicx}
\DeclareMathOperator*{\E}{\mathbb{E}}
\DeclareMathOperator{\R}{\mathbb{R}}
\DeclareMathOperator{\N}{\mathcal{N}}

\usepackage{hyperref}

\theoremstyle{definition}
\newtheorem{definition}{Definition}[section]
\newtheorem{propositions}{Proposition}
\newtheorem{corollary}{Corollary}[propositions]

\usepackage[accepted]{icml2019}

\icmltitlerunning{Neural Network Attributions: A Causal Perspective}

\begin{document}

\twocolumn[
\icmltitle{Neural Network Attributions: A Causal Perspective}



\icmlsetsymbol{equal}{*}

\begin{icmlauthorlist}
\icmlauthor{Aditya Chattopadhyay}{go}
\icmlauthor{Piyushi Manupriya}{to}
\icmlauthor{Anirban Sarkar}{to}
\icmlauthor{Vineeth N Balasubramanian}{to}
\end{icmlauthorlist}

\icmlaffiliation{go}{Center for Imaging Science, Johns Hopkins University, Baltimore, USA.}
\icmlaffiliation{to}{Department of Computer Science and Engineering, Indian Institute of Technology Hyderabad, Telangana, India}

\icmlcorrespondingauthor{Aditya Chattopadhyay}{achatto1@jhu.edu}
\icmlcorrespondingauthor{Vineeth N Balasubramanian}{vineethnb@iith.ac.in}


\vskip 0.3in
]



\printAffiliationsAndNotice{}  

\begin{abstract}
We propose a new attribution method for neural networks developed using first principles of causality (to the best of our knowledge, the first such). The neural network architecture is viewed as a Structural Causal Model, and a methodology to compute the causal effect of each feature on the output is presented. 
With reasonable assumptions on the causal structure of the input data, we propose algorithms to efficiently compute the causal effects, as well as scale the approach to data with large dimensionality. We also show how this method can be used for recurrent neural networks. We report experimental results on both simulated and real datasets showcasing the promise and usefulness of the proposed algorithm.

\end{abstract}
\vspace{-20pt}
\section{Introduction}
\label{introduction}
\vspace{-4pt}
Over the last decade, deep learning models have been highly successful in solving complex problems in various fields ranging from vision, speech to more core fields such as chemistry and physics \cite{deng2014deep,sadowski2014searching,gilmer2017neural}. However, a key bottleneck in accepting such models in real-life applications, especially risk-sensitive ones, is the ``interpretability problem". Usually, these models are treated as black boxes without any knowledge of their internal workings. This makes troubleshooting difficult in case of erroneous behaviour. Moreover, these algorithms are trained on a limited amount of data which most often is different from real-world data. Artifacts that creep into the training dataset due to human error or unwarranted correlations in data creation have an adverse effect on the hypothesis learned by these models. If treated as black boxes, there is no way of knowing whether the model actually learned a concept or a high accuracy was just fortuitous. This limitation of black-box deep learned models has paved way for a new paradigm, ``explainable machine learning".

While the field is nascent, several broad approaches have emerged \cite{simonyan2013deep, yosinski2015understanding, frosst2017distilling, letham2015interpretable}, each having its own perspective to explainable machine learning. In this work, we focus on a class of interpretability algorithms called ``attribution-based methods''. Formally, attributions are defined as the effect of an input feature on the prediction function's output \cite{sundararajan2017axiomatic}. This is an inherently causal question, which motivates this work. Current approaches involve backpropagating the signals to input to decipher input-output relations \cite{sundararajan2017axiomatic,selvaraju2016grad,bach2015pixel,ribeiro2016should} or approximating the local decision boundary (around the input data point in question) via ``interpretable" regressors like linear classifiers \cite{ribeiro2016should, selvaraju2016grad, zhou2015predicting, alvarez2017causal} or decision trees. 
 
In the former category of methods, while gradients answer the question ``How much would perturbing a particular input affect the output?'', they do not capture the causal influence of an input on a particular output neuron. The latter category of methods that rely on  ``interpretable" regression is also prone to artifacts as regression primarily maps correlations rather than causation. In this work, we propose a neural network attribution methodology built from first principles of causality. To the best of our knowledge, while neural networks have been modeled as causal graphs \cite{kocaoglu2017causalgan}, this is the first effort on a \textit{causal approach to attribution in neural networks}. 

Our approach views the neural network as a Structural Causal Model (SCM), and proposes a new method to compute the Average Causal Effect of an input neuron on an output neuron. Using standard principles of causality to make the problem tractable, this approach induces a setting where input neurons are not causally related to each other, but can be jointly caused by a latent confounder (say, data-generating mechanisms). This setting is valid in many application domains that use neural networks, including images where neighboring pixels are often affected jointly by a latent confounder, rather than direct causal influence (a ``doer'' can take a paint brush and oddly color a certain part of an image, and the neighboring pixels need not change). We first show our approach on a feedforward network, and then show how the proposed methodology can be extended to Recurrent Neural Networks which may violate this setting. We also propose an approximate computation strategy that makes our method viable for data with large dimensionality. We note that our work is different from a related subfield of structure learning \cite{eberhardt2007causation,hoyer2009nonlinear,hyttinen2013experiment,kocaoglu2017causalgan}, where the goal is to discern the causal structure in given data (for example, does feature $A$ cause feature $B$ or vice versa?). The objective of our work is to identify the causal influence of an input on a learned function's (neural network's) output.

Our key contributions can be summarized as follows. We propose a new methodology to compute causal attribution in neural networks from first principles; such an approach has not been expounded for neural network attribution so far to the best of our knowledge. We introduce causal regressors for better estimates of the causal effect in our methodology, as well as to provide a global perspective to causal effect. We provide a strategy to scale the proposed method to high-dimensional data. We show how the proposed method can be extended to Recurrent Neural Networks. We finally present empirical results to show the usefulness of this methodology, as well as compare it to a state-of-the-art gradient-based method to demonstrate its utility.


\section{Prior Work and Motivation}
\label{Section: previous_work}
\vspace{-4pt}
Attribution methods for explaining deep neural networks deal with identifying the effect of an input neuron on a specific output neuron. The last few years have seen a growth in research efforts in this direction \cite{sundararajan2017axiomatic,smilkov2017smoothgrad,shrikumar2017learning,montavon2017explaining,bach2015pixel}. Most such methods generate `saliency maps' conditioned on the given input data, where the map captures the contribution of a feature towards the overall function value. Initial attempts involved perturbing regions of the input via occlusion maps \cite{zeiler2014visualizing,zhou2015predicting} or inspecting the gradients of an output neuron with respect to an input neuron \cite{simonyan2013deep}. However, the non-identifiability of ``source of error'' has been a central impediment to designing attribution algorithms for black box deep models. It is impossible to distinguish whether an erroneous heatmap (given our domain knowledge) is an artifact of the attribution method or a consequence of poor representations learnt by the network \cite{sundararajan2017axiomatic}.

In order to analyze attribution methods in a uniform manner, newer methods \cite{sundararajan2017axiomatic} have spelt out axioms that can be used to evaluate a given method: (i) Conservativeness \cite{bach2015pixel}, (ii) Sensitivity, (iii) Implementation invariance, (iv) Symmetry preservation \cite{sundararajan2017axiomatic}, and (v) Input invariance \cite{kindermans2017reliability}. Methods that use the infinitesimal approximation of gradients and local perturbations violate axiom (ii). In flatter regions of the learned neural function, perturbing input features or investigating gradients might falsely point to zero attributions to these features. 

From a causal point of view, both gradient- and perturbation-based methods can be viewed as special instances of Individual Causal Effect (ICE), defined as, $ICE^y_{do(x_i = \alpha)} = y_{x_i = \alpha}(u) - y(u)$. $y_{x_i = \alpha}(u)$ denotes the output $y$ of the network for a given individual input vector $u$, with an arbitrary neuron $x_i$ set to $\alpha$. $y(u)$ represents the network output without any intervention. If input neurons are assumed to not cause each other, then calculating $ICE^y_{do(x_i = \alpha)}$ by setting $\alpha$ to $u_i + \epsilon$ can be related to taking the partial derivative, i.e., $\frac{\partial f}{\partial x_i}|_{x = u} = \frac{f(u_1, u_2, ..., u_i + \epsilon, .., u_n) - f(u_1, .., u_i, .., u_n)}{\epsilon} = \frac{y_{x_i = u_i + \epsilon}(u) - y(u)}{\epsilon} = \frac{ICE^y_{do(x_i = u_i + \alpha)}}{\epsilon}$ where $\epsilon \rightarrow 0$. Complex inter-feature interactions can conceal the real importance of input feature $x_i$, when only the ICE is analyzed. Appendix \ref{app_subsec_prior_work} provides more details of this observation.

Subsequent methods like DeepLIFT \cite{shrikumar2017learning} and LRP \cite{bach2015pixel} solved the sensitivity issue by defining an appropriate baseline and approximating the instantaneous gradients with discrete differences. This however, breaks axiom (iii), as unlike gradients, discrete gradients do not follow the chain rule \cite{shrikumar2017learning}. Integrated Gradients \cite{sundararajan2017axiomatic} extended this method to include actual gradients and averaged them out along a path from the baseline to the input vector. This method is perhaps closest to capturing causal influences since it satisfies most axioms among similar methods (and we use this for empirical comparisons in this work). Nevertheless, this method does not marginalize over other input neurons and the attributions may thus still be biased.

\vspace{-6pt}
\paragraph{Implicit biases in current attribution methods:} Kindermans \textit{et al.} \cite{kindermans2017reliability} showed that almost all attribution methods are sensitive to even a simple constant shift of all the input vectors. This implicitly means that the attributions generated for every input neuron are biased by the values of other input neurons for a particular input data. 
To further elucidate this point, consider a function $y = f(a,b) = ab$. Let the baseline be $[a_{base},b_{base}] = [2,2]$. Consider two input vectors $[3,5]$ and $[3,100]$. The Integrated Gradients method (which unlike other methods, satisfies all the axioms in Section \ref{Section: previous_work} except axiom (v)) assigns attributions to $[a,b]$ as $[3.4985, 7.4985]$ for input $[3,5]$ and $[50.951, 244.951]$ for input $[3,100]$. This result is misleading, because both input vectors have exactly the same baseline and same value for feature $a = 3$, but the attribution algorithm assigns different values to it. However, because the form of the function is known \textit{a priori}, it is clear that both $a$ and $b$ have equal causal strengths towards affecting $y$, and in this particular scenario, the entire change in $y$ is due to interventions on $b$ and not $a$.

In this work, we propose a causal approach to attribution, which helps supersede the implicit biases in current methods by marginalizing over all other input parameters. We show in Section \ref{Section: ACE_method}, after our definitions, that our approach to causal attribution satisfies all axioms, with the exception of axiom (i), which is not relevant in a causal setting. Besides, via the use of causal regressors \ref{section: Calculating the baseline values}, a global perspective of the deep model can be obtained, which is not possible by any existing attribution method.


The work closest to ours is a recent effort to use causality to explain deep networks in natural language processing \cite{alvarez2017causal}. This work is a generalization of LIME \cite{ribeiro2016should}, where the idea is to infer dependencies via regularized linear regression using perturbed samples local to a particular input. Analyzing the weights of this learned function provides insights into the network's local behavior. However, regression only learns correlations in data which could be markedly different from causation. Other efforts such as \cite{alvarez2018towards, li2018deep} attempt to explain in terms of latent concepts, which again do not view effect from a causal perspective, which is the focus of this work. More discussion of prior work is presented in Appendix \ref{app_subsec_prior_work}.

\vspace{-4pt}
\section{Background: Neural Networks as Structural Causal Models (SCMs)}
\label{Section: philosophy}
\vspace{-4pt}
This work is founded on principles of causality, in particular Structural Causal Models (SCMs) and the $do(.)$ calculus, as in \cite{pearl2009causality}. A brief exposition on the concepts used in this work is provided in Appendix \ref{Basics of causality}. 

We begin by stating that neural network architectures can be trivially interpreted as SCMs (as shown in other recent work such as \cite{kocaoglu2017causalgan}). 
Note that we do not explicitly attempt to find the causal direction in this case, but only identify the causal relationships given a learned function.
%
%
\begin{figure}[h]
  \centering 
  \includegraphics[scale=0.5]{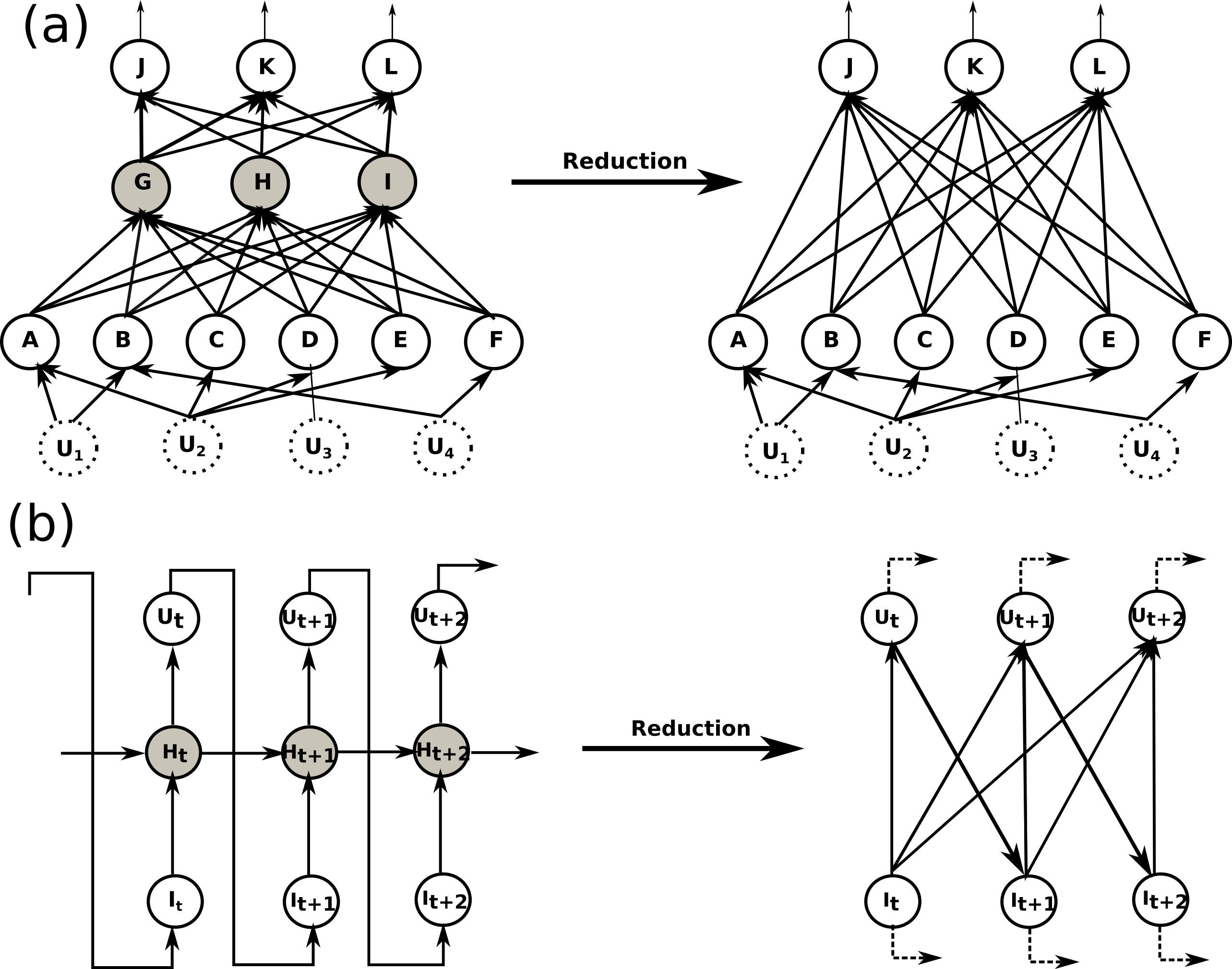}
  \caption{(a) Feedforward neural network as an SCM. The dotted circles represent exogenuous random variables which can serve as common causes for different input features. (b) Recurrent neural network as an SCM.}
  \vspace{-5pt}
  \label{neural_network}
\end{figure}
Figure \ref{neural_network}a depicts such a feedforward neural network architecture. Neural networks can be interpreted as directed acyclic graphs with directed edges from a lower layer to the layer above. The final output is thus based on a hierarchy of interactions between lower level nodes. 

\begin{propositions}
An $l$-layer feedforward neural network $N (l_1, l_2, ...l_n)$ where $l_i$ is the set of neurons in layer $i$ has a corresponding SCM $M([l_1, l_2, ...., l_n], U, [f_1, f_2, ... f_n], P_U)$, where $l_1$ is the input layer and $l_n$ is the output layer. Corresponding to every $l_i$, $f_i$ refers to the set of causal functions for neurons in layer $i$. $U$ refers to a set of exogenous random variables which act as causal factors for the input neurons $l_1$. 
\label{prop: NN as SCM}
\end{propositions}

Appendix \ref{appendix: proposition 2} contains a simple proof of Proposition \ref{prop: NN as SCM}. In practice, only the neurons in layer $l_1$ and layer $l_n$ are observables, which are derived from training data as inputs and outputs respectively. The causal structure can hence be reduced to SCM $M([l_1, l_n], U, f', P_U)$ by marginalizing out the hidden neurons. 

\begin{corollary}
Every $l$-layer feedforward neural network $N (l_1, l_2, ...l_n)$, with $l_i$ denoting the set of neurons in layer $i$, has a corresponding SCM $M([l_1, l_2, ...., l_n], U, [f_1, f_2, ... f_n], P_u)$ which can be reduced to an SCM $M'([l_1, l_n], U, f', P_U)$. 
\label{corr: recursive_subs}
\end{corollary}

Appendix \ref{appendix: corollary 2} contains a formal proof for Corollary \ref{corr: recursive_subs}. Marginalizing the hidden neurons out by recursive substitution (Corollary \ref{corr: recursive_subs}) is analogous to deleting the edges connecting these nodes and creating new directed edges from the parents of the deleted neurons to their respective child vertices (the neurons in the output layer) in the corresponding causal Bayesian network. Figure \ref{neural_network}a illustrates an example of a 3-layer neural network (the left figure) with 1 input, 1 hidden and 1 output layer (W.l.o.g); after marginalizing out the hidden layer neurons, the reduced causal Bayesian network on the right is obtained.

\vspace{-4pt}
\paragraph{Recurrent Neural Networks (RNNs):}
Defining an SCM directly on a more complex neural network architecture such as RNNs would introduce feedback loops and the corresponding causal Bayesian network is no longer acyclic. Cyclic SCMs may be ambiguous and not register a unique probability distribution over its endogenous variables \cite{bongers2016structural}.  Proposition \ref{prop: NN as SCM}, however, holds for a time-unfolded RNN; but care must be taken in defining the reduced SCM $M'$ from the original SCM $M$. Due to the recurrent connections between hidden states, marginalizing over the hidden neurons (via recursive substitution) creates directed edges from input neurons at every timestep to output neurons at subsequent timesteps. In tasks such as sequence prediction, where the output neuron $U_t$ at time $t$ is taken as the input at time $t+1$, the assumption that input neurons are not causally related is violated. We discuss this in detail in Section \ref{subsec_causal_attrib_RNNs}.
Figure \ref{neural_network}b depicts our marginalization process in RNNs. W.l.o.g., we consider a single hidden layer unfolded recurrent model where the outputs are used as inputs for the next time step. The shaded vertices are the hidden layer random variables, $U_i$ refers to the output at time $i$ and $I_i$ refers to the input at time $i$.
In the original SCM $M$ (left figure), vertex $H_{t+1}$ causes $U_{t+1}$ (there exists a functional dependence). If $H_{t+1}$ is marginalized out, its parents $I_{t+1}$ and $H_t$ become the causes (parents) of $U_{t+1}$. Similarly, if $H_t$ is marginalized out, both $I_t$ and $I_{t+1}$ become causes of $U_{t+1}$. Using similar reasoning, the reduced (marginalized) SCM $M'$ on the right is obtained. 

\vspace{-4pt}
\section{Causal Attributions for Neural Networks}
\label{Section: ACE_method}
\vspace{-4pt}
\subsection{Causal Attributions}
\label{subsec_causal_attrib}
\vspace{-4pt}
This work attempts to address the question: 
"What is the causal effect of a particular input neuron on a particular output neuron of the network?". This is also known in literature as the ``attribution problem" \cite{sundararajan2017axiomatic}. We seek the information required to answer this question as encapsulated in the SCM $M'([l_1 , l_n], U, f', P_U)$ consistent with the neural model architecture $N (l_1, l_2, ...l_n)$. 

\theoremstyle{definition}
\begin{definition}({Average Causal Effect}).
The \textit{Average Causal Effect (ACE)} of a binary random variable $x$ on another random variable $y$ is commonly defined as $\E[y|do(x = 1)] - \E[y|do(x = 0)]$.
\label{def: ACE}
\end{definition}

While the above definition is for binary-valued random variables, the domain of the function learnt by neural networks is usually continuous. Given a neural network with input $l_1$ and output $l_n$, we hence measure the $ACE$ of an input feature $x_i \in l_1$ with value $\alpha$ on an output feature $y \in l_n$ as: 
\begin{equation}
ACE^y_{do(x_i = \alpha)} = \E[y|do(x_i = \alpha)] - baseline_{x_i}
\label{eq: ACE}
\end{equation}

\theoremstyle{definition}
\begin{definition}({Causal Attribution}).
We define $ACE^y_{do(x_i = \alpha)}$ as the \textit{causal attribution} of input neuron $x_i$ for an output neuron $y$.
\label{def: causal attribution}
\end{definition}

Note that the gradient $\frac{\partial \E[y|do(x_i = \alpha)]}{\partial x_i}$ is sometimes used to approximate the Average Causal Effect ($ACE$) when the domain is continuous \cite{peters2017elements}. However, as mentioned earlier, gradients suffer from sensitivity and induce causal effects biased by other input features. Also, it is trivial to see that our definition of causal attributions satisfy axioms (ii) - (vi) (as in Section \ref{Section: previous_work}), with the exception of axiom (i). According to axiom (i), $atr$ is conservative if $\sum_i atr_i = f(inp) - f(baseline)$, where $atr$ is a vector of attributions for the input. However, our method identifies the causal strength of various input features towards a particular output neuron and not a linear approximation of a deep network, so it's not necessary for the causal effects to add up to the difference between $f(inp)$ and $f(baseline)$. Axiom (ii) is satisfied due to the consideration of a reference baseline value. Axioms (iii) and (iv) hold because we directly calculate the interventional expectations which do not depend on the implementation as long as it maps to an equivalence function.  \cite{kindermans2017reliability} show that most attribution algorithms are very sensitive to constant shifts in the input. In the proposed method, if two functions $f_1(x) = f_2(x + c)$ $\forall x$, where $c$ is the constant shift, the respective causal attributions of $x$ and $x+c$ stay exactly the same. Thus, our method also satisfies axiom (v).


In Equation \ref{eq: ACE}, an ideal baseline would be any point along the decision boundary of the neural network, where predictions are neutral. However, \cite{kindermans2017reliability} showed that when a reference baseline is fixed to a specific value (such as a zero vector), attribution methods are not affine-invariant. In this work, we propose the average ACE of $x_i$ on $y$ as the baseline value for $x_i$, i.e. $baseline_{x_i} = \E_{x_i}[\E_y[y|do(x_i = \alpha)]]$. In absence of any prior information, we assume that the ``doer" is equally likely to perturb $x_i$ to any value between $[low^i, high^i]$, i.e. $x_i \sim U(low^i, high^i)$, where $[low^i, high^i]$ is the domain of $x_i$. While we use the uniform distribution, which represents the maximum entropy distribution among all continuous distributions in a given interval, if more information about the distribution of interventions performed by the ``external" doer is known, this could be incorporated instead of an uniform distribution. Domain knowledge could also be incorporated to select a significant point $\hat{x_i}$ as the baseline. The $ACE^y_{do(x_i = \alpha)}$ would then be $\E[y|do(x_i = \alpha)] - \E[y|do(x_i=\hat{x_i})]$.
Our choice of baseline in this work is unbiased and adaptive. Another rationale behind this choice is that $\E[y|do(x_i = \alpha)]$ represents the expected value of random variable $y$ when the random variable $x_i$ is set to $\alpha$. If the expected value of $y$ is constant for all possible interventional values of $x_i$, then the causal effect of $x_i$ on $y$ would be $0$ for any value of $x_i$. The baseline value in that case would also be the same constant, resulting in $ACE^y_{do(x_i = \alpha)} = 0$. 
\vspace{-4pt}
\subsection{Calculating Interventional Expectations}
\label{section: Calculating the interventional expectations}
\vspace{-4pt}
We refer to $\E[y|do(x_i = \alpha)]$ as the \textit{interventional expectation} of $y$ given the intervention $do(x_i = \alpha)$. By definition:
\vspace{-5pt}
\begin{equation}
\E[y|do(x_i = \alpha)] = \int_{y} y p(y|do(x_i = \alpha)) dy 
\label{interventional eq}
\end{equation}
\vspace{-10pt}

Naively, evaluating Equation \ref{interventional eq} would involve sampling all other input features from the empirical distribution keeping feature $x_i = \alpha$, and then averaging the output values. Note, this assumes that the input features don't cause one another. However, due to the curse of dimensionality, this unbiased estimate of $\E[y|do(x_i = \alpha)]$ would have a high variance. Moreover, running through the entire training data for each interventional query would be time-consuming. We hence propose an alternative mechanism to compute the interventional expectations.

Consider an output neuron $y$ in the reduced SCM $M'([l_1, l_n], U, f', P_U)$, obtained by marginalizing out the hidden neurons in a given neural network $N(l_1, l_2,....l_n)$ (Corollary \ref{corr: recursive_subs}). The causal mechanism can be written as $y = f'_y(x_1, x_2, ..., x_k)$, where $x_i$ refers to neuron $i$ in the input layer, and $k$ is the number of input neurons. If we perform a $do(x_i = \alpha)$ operation on the network, the  causal mechanism is given by $y = f'_{y|do(x_i = \alpha)}(x_1, ..., x_{i - 1}, \alpha, x_{i + 1}, ..., x_k)$. For brevity, we drop the $do(x_i = \alpha)$ subscript and simply refer to this as $f'_y$. Let $\mu _j = \E[x_j|do(x_i = \alpha)] \forall x_j \in l_1$. Since $f'_y$ is a neural network, it is smooth (assuming smooth activation functions). Now, the second-order Taylor's expansion of the causal mechanism $f'_{y|do(x_i = \alpha)}$ around the vector $\mu = [\mu_1, \mu_2, ..., \mu_k]^T$ is given by (recall $l_1$ is the vector of input neurons):
\vspace{-5pt}
\begin{multline}
f'_{y}(l_1) \approx f'_y(\mu)  + \nabla^Tf'_y(\mu)(l_1 - \mu) + \\
\frac{1}{2}(l_1 - \mu)^T\nabla^2f'_y(\mu)(l_1 - \mu)
\label{taylor expansion eq}
\end{multline}
Taking expectation on both sides (marginalizing over all other input neurons): 
\vspace{-5pt}
\begin{multline}
\E[f'_{y}(l_1)| do(x_i = \alpha)] \approx f'_y(\mu) + \\
\frac{1}{2}Tr(\nabla^2f'_y(\mu)\E[(l_1 - \mu)(l_1 - \mu)^T| do(x_i = \alpha)])
\vspace{-4pt}
\label{expected taylor expansion eq}
\end{multline}
The first-order terms vanish because $\E(l_1|x_i = \alpha) = \mu$. We now only need to calculate the individual interventional means $\mu$ and, the interventional covariance between input features $\E[(l_1 - \mu)(l_1 - \mu)^T| do(x_i = \alpha)]$ to compute Equation \ref{interventional eq}. 
Such approximations of deep non-linear neural networks via Taylor's expansion have been explored before in the context of explainability \cite{montavon2017explaining}, though their overall goal was different. 

While every SCM $M'$, obtained via marginalizing out the hidden neurons, registers a causal Bayesian network, this network is not necessarily causally sufficient (Reichenbach's common cause principle) \cite{pearl2009causality}. There may exist latent factors or noise which jointly cause the input features, i.e., the input features need not be independent of each other. We hence propose the following.

\begin{propositions}
Given an $l$-layer feedforward neural network $N (l_1, l_2, ...l_n)$ with $l_i$ denoting the set of neurons in layer $i$ and its corresponding reduced SCM $M'([l_1, l_n], U, f', P_U)$, the intervened input neuron is d-separated from all other input neurons.
\label{prop: d-sep feedforward}
\end{propositions}

Appendix \ref{appendix: proposition 3} provides the proof for Proposition \ref{prop: d-sep feedforward}.

\begin{corollary}
Given an $l$-layer feedforward neural network $N (l_1, l_2, ...l_n)$ with $l_i$ denoting the set of neurons in layer $i$ and an intervention on neuron $x_i$, the probability distribution of all other input neurons does not change, i.e. $\forall x_j \in l_1$ and $x_j \neq x_i$ $P(x_j|do(x_i = \alpha)) = P(x_j)$.
\label{corr: independent nodes}
\end{corollary}

The proof of Corollary \ref{corr: independent nodes} is rather trivial and directly follows from Proposition \ref{prop: d-sep feedforward} and d-seperation \cite{pearl2009causality}. Thus, the interventional means and covariances are equal to the observational means and covariances respectively. The only intricacy involved now is in the means and covariances related to the intervened input neuron $x_i$. Since $do(x_i = \alpha)$, these can be computed as $\E[x_i|do(x_i = \alpha)] = \alpha$ and $Cov(x_i, x_j|do(x_i = \alpha)) = 0$ $\forall x_j \in l_1$ (the input layer).

In other words, Proposition \ref{prop: d-sep feedforward} and Corollary \ref{corr: independent nodes} induce a setting where causal dependencies (functions) do not exist between different input neurons. This assumption is often made in machine learning models (where methods like Principal Component Analysis are applied if required to remove any correlation between the input dimensions). If there was a dependence between input neurons, that is due to latent confounding factors (nature) and not the causal effect of one input on the other. Our work is situated in this setting. 
This assumption is however violated in the case of time-series models or sequence prediction tasks, which we handle later in Section \ref{subsec_causal_attrib_RNNs}. 
\vspace{-4pt}
\subsection{Computing ACE using Causal Regressors}
\label{section: Calculating the baseline values}
\vspace{-4pt}
The ACE (Eqn \ref{eq: ACE}) requires the computation of two quantities: the interventional expectation and the baseline. We defined the baseline value for each input neuron to be $\E_{x_i}[\E_y[y|do(x_i = \alpha)]]$. In practice, we evaluate the baseline by perturbing the input neuron $x_i$ uniformly in fixed intervals from [$low^i, high^i$], and computing the interventional expectation.

The interventional expectation $\E[y|do(x_i = \alpha)]$ is a function of $x_i$ as all other variables are marginalized out. In our implementations, we assume this function to be a member of the polynomial class of functions $\{f | f(x_i) = \Sigma_{j}^{order}w_jx_i^j\}$ (this worked well for our empirical studies, but can be replaced by other classes of functions if required). Bayesian model selection \cite{claeskens2008model} is employed to determine the optimal order of the polynomial that best fits the given data by maximizing the marginal likelihood. The prior in Bayesian techniques guard against overfitting in higher order polynomials. $\E_{x_i}[\E_y[y|do(x_i = \alpha)]]$ can then be easily computed via analytic integration using the predictive mean as the coefficients of the learned polynomial model. The predictive variance of $y$ at any point $do(x_i = \alpha)$ gives an estimate of the model's confidence in its decision. If the variance is too high, more sampling of the interventional expectation at different $\alpha$ values may be required. For more details, we urge interested readers to refer to \cite{christopher2016pattern}[Chap 3]. We name the learned polynomial functions \textit{\textbf{causal regressors}}. $ACE^y_{do(x_i = \alpha)}$ can thus be obtained by evaluating the causal regressor at $x_i = \alpha$ and subtracting this value from the $baseline_{x_i}$. Calculating interventional expectations for multiple input values is a costly operation; learning causal regressors allows one to estimate these values on-the-fly for subsequent attribution analysis. Note that other regression techniques like spline regression can also be employed to learn the interventional expectations. In this work, the polynomial class of functions was selected for its mathematical simplicity. 

\vspace{-4pt}
\subsection{Overall Methodology}
\label{method}
\vspace{-4pt}
We now summarize our overall methodology to compute causal attributions of a given input neuron for a particular output neuron in a feedforward neural network (Defn \ref{def: causal attribution}). Phase I of our method computes the interventional expectations (Sec \ref{section: Calculating the interventional expectations}) and Phase II learns the causal regressors and estimates the baseline (Sec \ref{section: Calculating the baseline values}).
\vspace{-9pt}
\paragraph{Phase I:}
For \textit{feedforward networks}, the calculation of interventional expectations is straightforward. The empirical means and covariances between input neurons can be precomputed from training data (Corollary \ref{corr: independent nodes}). Eqn \ref{expected taylor expansion eq} is computed using these empirical estimates to obtain the interventional expectations, $\E[y|do(x_i = \alpha)]$, for different values of $\alpha$. Appendix \ref{Algorithm for Phase 1 in case of Feedforward networks} presents a detailed algorithm/pseudocode along with its complexity analysis. In short, for $num$ different interventional values and $k$ input neurons, the algorithmic complexity of Phase I for feedforward networks would be O($num \times k$).
\vspace{-9pt}
\paragraph{Phase II:}
As highlighted earlier, calculating interventional expectations can be costly; so, we learn a causal regressor function that can approximate this expectation for subsequent on-the-fly computation of interventional expectations. The output of Phase I (interventional expectations at $num$ different interventions on $x_i$) is used as training data for the polynomial class of functions (Sec \ref{section: Calculating the baseline values}). The causal regressors are learned using Bayesian linear regression, and the learned model is used to provide the interventional expectations for out-of-sample interventions. Appendix \ref{phase II algorithm} presents a detailed algorithm.
\vspace{-4pt}
\subsection{Causal Attribution in RNNs}
\label{subsec_causal_attrib_RNNs}
\vspace{-4pt}
As mentioned before, the setting where causal dependencies do not exist between different input neurons is violated in the case of RNNs. 
In the corresponding causally sufficient Bayesian network $G^c = (V,E)$ for a recurrent architecture, 
input neurons $\{I_{t+1}, I_{t+2}\}$ are not independent from $I_t$ after an intervention on $I_t$ as they are d-connected \cite{pearl2009causality} (see Figure \ref{neural_network}b). 
For a \textit{recurrent neural network}(RNN), if it does not have output to input connections, then the unfolded network can be given the same treatment as feedforward networks for calculating $\E[y|do(x_i = \alpha)]$. However, in the presence of recurrent connections from output to input layers, the probability distribution of the input neurons at subsequent timesteps would change after an intervention on neuron $x_i^{\hat{t}}$ ($i^{th}$ input feature at time $\hat{t}$). As a result, we cannot precompute the empirical covariance and means for use in Equation \ref{expected taylor expansion eq}. In such a scenario, means and covariances are estimated after evaluating the RNN over each input sequence in the training data with the value at $x_i^{\hat{t}} = \alpha$. This ensures that these empirical estimates are calculated from the interventional distribution $P(. | do(x^{\hat{t}}_i = \alpha))$. Eqn \ref{expected taylor expansion eq} is then evaluated to obtain the interventional expectations. Appendix \ref{Algorithm for Phase 1 in case of Recurrent networks} presents a detailed algorithm/pseudocode. The complexity per input neuron $x_i^{\hat{t}}$ is O($n \times num$), with $n$ training samples and $num$ interventional values. The overall complexity scales linearly with the timelag $\tau$ for causal attributions for a particular output $y$ at timestep $t$.
\vspace{-4pt}
\begin{propositions}
Given a recurrent neural function, unfolded in the temporal dimension, the output at time $t$ will be ``strongly" dependent on inputs from timesteps $t$ to $t - \tau$, where $\tau \triangleq \E_x[\max_k(|det(\nabla_{x^{t - k}}y^t)| > 0)]$.
\label{proposition: get tau}
\end{propositions}
\vspace{-4pt}
We present the proof for Proposition \ref{proposition: get tau} in Appendix \ref{proof proposition 4}. $\tau$ can be easily computed per sample with a single backward pass over the computational graph. This reduces the complexity of understanding causal attributions of all features for a particular output at time $t$ from O(n.num.t.k) to O(n.num.$\tau$.k). Here $k$ is the number of input neurons at each time-step.
\vspace{-4pt}
\subsection{Scaling to Large Data} 
\vspace{-4pt}
Evaluating the interventional expectations using Eqn \ref{expected taylor expansion eq} involves calculating the Hessian. 
Note however that we never explicitly require the Hessian, just the term $\sum_{i=1}^k\sum_{j=1}^k\nabla^2f'_y(\mu)_{ij}Cov(x_i, x_j|do(x_l = \alpha))$. We provide an efficient methodology to compute the interventional expectations for high-dimensional data, using the Taylor series expansion of $f'_y$ around $\mu$ and the eigendecomposition of $Cov(\mathbf{x},\mathbf{x}|do(x_l = \alpha)) = \sum_{r=1}^k\lambda_r e_r e_r^T$. This allowed us to get results significantly faster than exact calculations (0.04s for the approximation v/s 3.04s per computation for experiments on MNIST dataset with a deep neural network of 4 hidden layers). More details are provided in Appendix \ref{section: scalability}.

\vspace{-4pt}
\section{Experiments and Results}
\label{Section: experiments}
The implementation of our method is publicly available at \href{https://github.com/Piyushi-0/ACE}{https://github.com/Piyushi-0/ACE}.
\vspace{-4pt}
\subsection{Iris dataset}
\begin{figure}[t]
\centering
\includegraphics[height=5.5cm, width=8.5cm]
{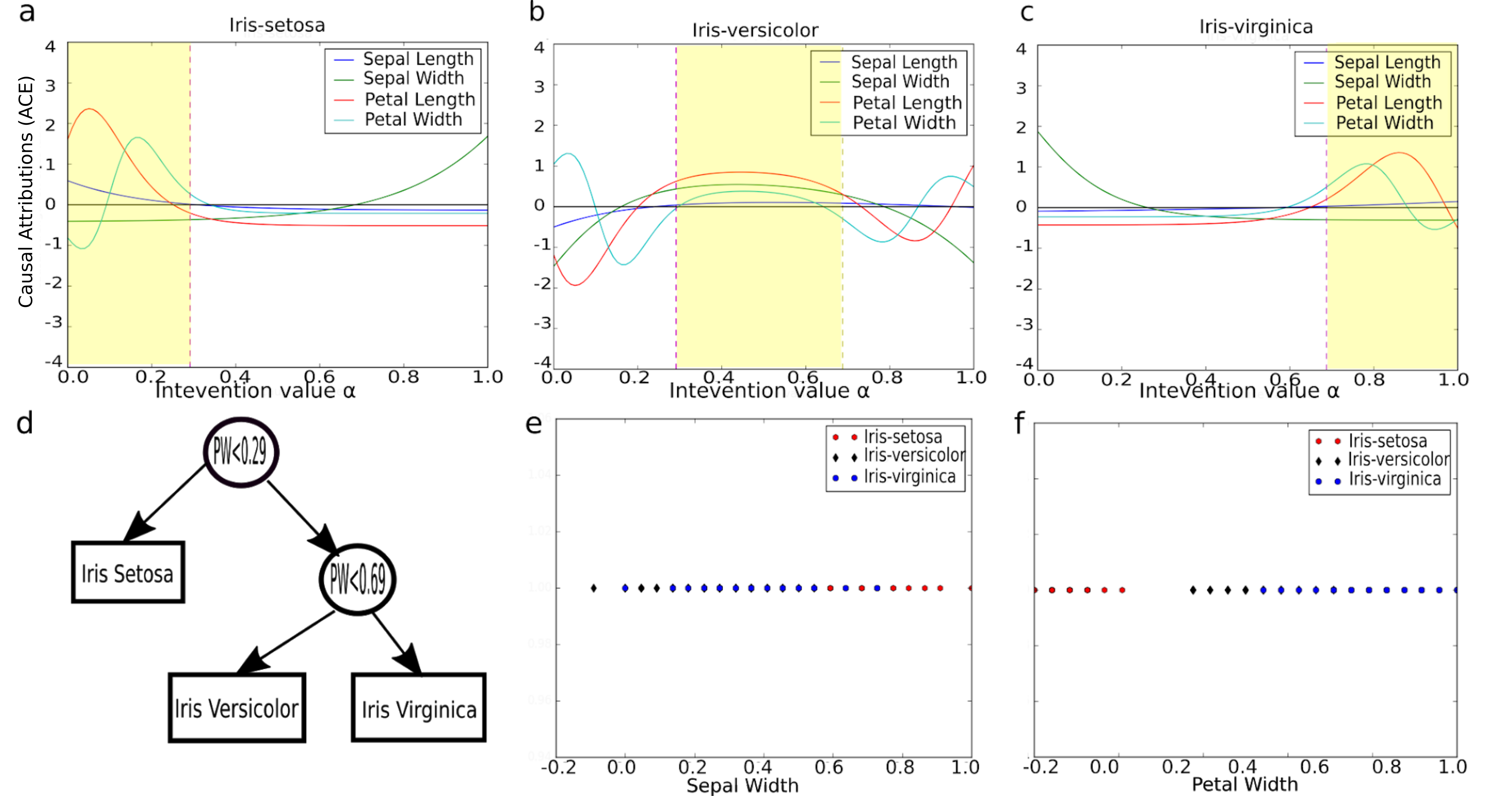}
\vspace{-5pt}
  \caption{Results for the proposed method on the Iris dataset. a,b,c) causal regressors for Iris-setosa, Iris-versicolor \& Iris-virginica respectively; d) decision tree trained on Iris dataset; e,f) scatter plots for sepal and petal width for all three Iris dataset classes. (Best viewed in color)}
  \label{iris_results}
  \vskip -0.15in
\end{figure}

\vspace{-4pt}
A 3-layer neural network (with  \textit{relu()} activation functions) was trained on the Iris dataset \cite{Dua:2017}. All the input features were [0-1] normalized. Fig \ref{iris_results} shows how our method provides a powerful tool for deciphering neural decisions at an individual feature level. Figs \ref{iris_results} a, b \& c depict causal regressors for the three classes and all four features. These plots easily reveal that smaller petal length and width are positively causal ($ACE \geq 0$) for Iris-setosa class; moderate values can be attributed to Iris-versicolor; and higher values favor the neural decision towards Iris-verginica. Due to the simplicity of the data, it can be almost accurately separated with axis-aligned decision boundaries. Fig \ref{iris_results}d, shows the structure of the learned decision tree. PW refers to the feature petal width. The yellow colored sections in Figs \ref{iris_results}a, b and c are the regions where the decision tree predicts the corresponding class by thresholding the petal width value. In all three figures, the causal regressors show strong positive ACE of petal width for the respective classes. Figs \ref{iris_results} e and f are scatter plots for sepal width and petal width respectively for all the three classes. Figure \ref{iris_results}f clearly shows that $PW_{virginica} > PW_{versicolor} > PW_{setosa}$ (in accordance with the inference from Figs \ref{iris_results}a, b and c). Interestingly, the trend is reversed for sepal width, which has also been identified by the neural network as evident from Figs \ref{iris_results}a and c. Note that such a global perspective on explaining neural networks is not possible with any other attribution method.
\vspace{-5pt}
\subsection{Simulated data}
\vspace{-4pt}
\label{subsection: simulated_data}
\begin{figure}
  \centering
  \includegraphics[scale=0.125]{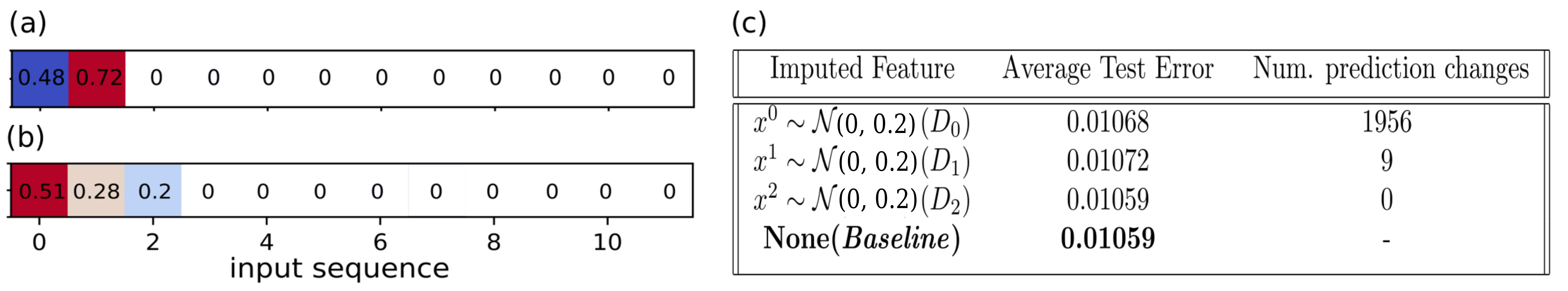}
  \vspace{-6pt}
  \caption{Saliency maps on test using (a) Causal attributions; (b) Integrated Gradients; (c) Imputation experiments (Sec \ref{subsection: simulated_data}). \textit{Num. prediction changes} were evaluated over 1M test sequences.}
  \label{LSTM toy exp}
  \vskip -0.2in
\end{figure}
Our approach can also help in generating local attributions just like other contemporary attribution algorithms. Causal attributions of each input neuron $x$ for output $y$ with $ACE^y_{do(x = input[x])}$  ($input[x]$ refers to the input vector value at neuron $x$), can be used as a saliency map to explain the local decisions. The simulated dataset is generated following a similar procedure used in the original LSTM paper \cite{hochreiter1997long} (procedure described in Appendix \ref{generation of simulated dataset}). Only the first three features of a long sequence is relevant for the class label of that sequence. A Gated Recurrent Unit (GRU) with a single input, hidden and output neuron with \textit{sigmoid()} activations is used to learn the pattern. The trained network achieves an accuracy of $98.94\%$. We compared the saliency maps generated by our method with Integrated Gradients (IG) \cite{sundararajan2017axiomatic} because it is the only attribution method that satisfies all the axioms, except axiom (v) (Section \ref{Section: previous_work}). The saliency maps were thresholded to depict only positive contributions. Figures \ref{LSTM toy exp}a and b show the results. 

By construction, the true recurrent function should consider only the first three features as causal for class prediction. While both IG and causal attributions associate positive values to the first two features, a $0$ attribution for the third feature (in Fig \ref{LSTM toy exp}a) might seem like an error of the proposed method.
A closer inspection however reveals that the GRU does not even look at the third feature before assigning a label to a sequence. From the simulated test dataset, we created three separate datasets $D_i$ by imputing the $i^{th}$ feature as $x^i \sim \mathcal{N}(0,0.2)$, $0 \leq i < 3$. Each $D_i$ was then passed through the GRU and the average test error was calculated. The results in Fig \ref{LSTM toy exp}c indicate that the third feature was never considered by the learned model for classifying the input patterns. While imputing $x^0$ and $x^1$ changed the LSTM's prediction $1956$ and $9$ times respectively, when evaluated over 1M sequences, imputing $x^3$ had no effect. IG heatmaps (Fig \ref{LSTM toy exp}b) did not detect this due to biases induced by strong correlations between input features.

\vspace{-5pt}
\subsection{Airplane Data}
\vspace{-4pt}
\begin{figure}
\begin{center}
\centerline{\includegraphics[width=0.98\columnwidth]{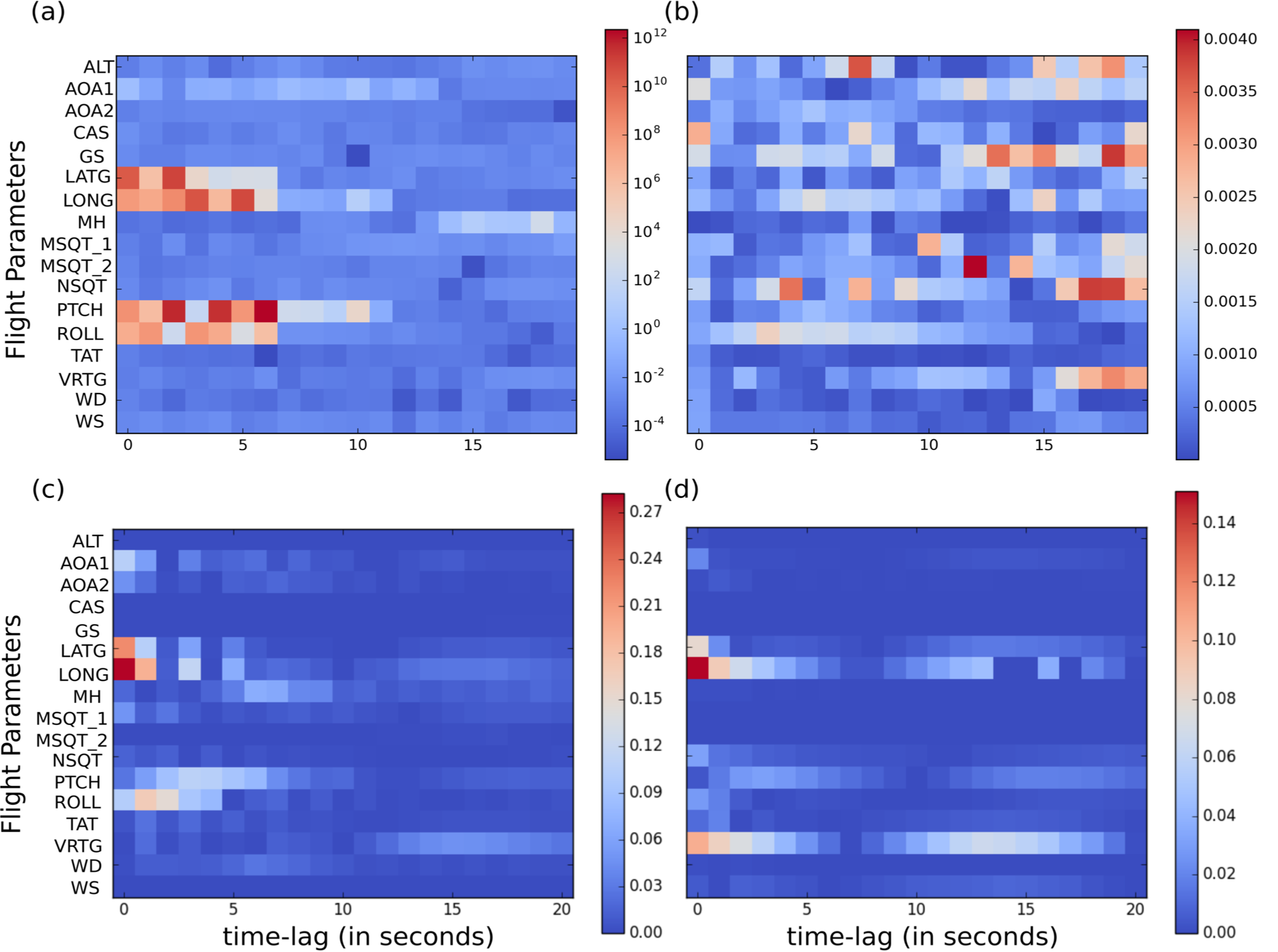}}
\caption{Causal attributions for (a) an anomalous flight and (b) a normal flight. IG attributions for the same (c) anomalous flight and (d) normal flight. All saliency maps are for the LATG parameters 60 seconds after touchdown.}
  \label{aircraft data}
  \vskip -0.4in
\end{center}
\end{figure}
We used a publicly available NASA Dashlink flight dataset (https://c3.nasa.gov/dashlink/projects/85/) to train a single hidden layer LSTM. The LSTM learns the flight's trajectory, with outputs used as inputs in the next timestep. The optimal lag-time was determined to be $\sim 20s$ (Proposition \ref{proposition: get tau}). Given a flight trajectory, to compute $ACE_{do(x_i^{\hat{t}} = \alpha)}^{y^t}$, we intervene on the LSTM by simulating the trajectory with $x_i^{\hat{t}}=\alpha$ for all trajectories in the train set (all input features $t < {\hat{t}}$ are taken from train set). The interventional means and covariances are then computed from these simulated trajectories and used in Eqn \ref{expected taylor expansion eq} (See Algorithm \ref{algo: phase I recurrent} in the Appendix). Fig \ref{aircraft data}a depicts the results for a specific flight, which was deemed as an anomaly by the Flight Data Recorder (FDR) report (due to slippery runway, the pilot could not apply timely brakes, resulting in a steep acceleration in the airplane post-touchdown). Observing the causal attributions for the lateral acceleration (LATG) parameter $60$ seconds post-touchdown shows strong causal effects in the Lateral acceleration (LATG), Longitudinal acceleration (LONG), Pitch (PTCH) and Roll (ROLL) parameters of the flight sequence up to 7 seconds before. These results strongly agree with the FDR report. For comparison, Fig \ref{aircraft data}b shows the causal attributions for a normal flight which shows no specific structure in its saliency maps. Figs \ref{aircraft data} c and d show explanations generated for the same two flights using the IG method. Unlike causal attributions, a stark difference in the right and left saliency maps is not visible. 
\vspace{-4pt}
\subsection{Visualizing Causal Effect}
\label{section: disentangled_latent_representations}
\vspace{-4pt}
\begin{figure}
\begin{center}
\centerline{\includegraphics[width=0.95\columnwidth]{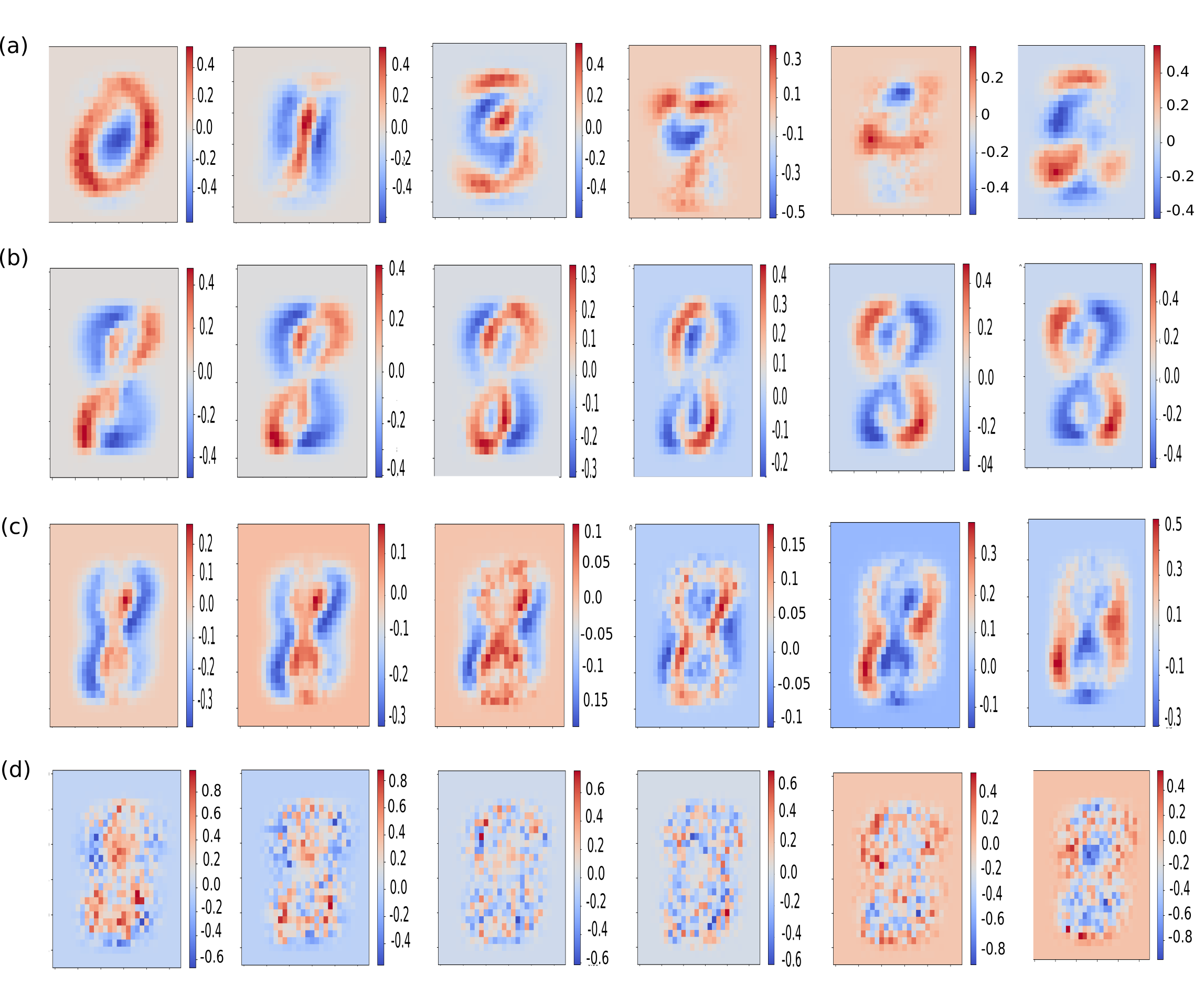}}
\caption{Causal attributions of (a) $c_k$ (class-specific latents), (b) $z_0$ \& $c_8$, (c) $z_6$ \& $c_8$, (d) $z_2$ \& $c_8$ for decoded image (Sec \ref{section: disentangled_latent_representations})}
  \label{decoder images}
  \vskip -0.45in
\end{center}
\end{figure}

In order to further study the correctness of our causal attributions, we evaluated our algorithm on data where explicit causal relations are known. 
In particular, if a dimension in the representation represents unique generative factors, they can be regarded as causal factors for data. To this end, we train a conditional \cite{kingma2014semi} $\beta$-VAE  \cite{higgins2016beta} on MNIST data to obtain disentangled representations which represent unique generative factors. 
The latent variables were modeled as 10 discrete variables (for each digit class) $[c_0, c_1, ..., c_9]$ (which were conditioned on while training the VAE) and 10 continuous variables (for variations in the digit such as rotation and scaling) $[z_0, z_1, z_2, ..., z_9]$.  $\beta$ was set to $10$. Upon training, the generative decoder was taken and $ACE^{x_{ij}}_{do(z_k = \alpha), do(c_l = 1)}$ and $ACE^{x_{ij}}_{do(c_k = \alpha)}$ (Defn. \ref{def: causal attribution}) were computed for each decoded pixel $x_{ij}$ and intervened latent variables $c_k/c_l/z_k$ $\forall k,l \in {0, 1, ..., 9}$. In case of continuous latents, along with each $z_k$, $c_l$ is also intervened on (ensuring $\sum_{l=0}^9c_l = 1$) to maintain consistency with the generative process. Since we have access to a probabilistic model through the VAE, the interventional expectations were calculated directly via Eqn \ref{interventional eq}. For each $z_k$, the baseline was computed as in Sec \ref{subsec_causal_attrib}. For the binary $c_k$s, we took $\E[x_{ij}|do(c_k = 0)]$ as the baseline. (More details are in Appendix \ref{decoder_details}.)

Fig \ref{decoder images}a corresponds to ACE of $c_0, c_1, c_3, c_7, c_4, c_2$ (from left to right) on each pixel of the decoded image (as output). The results indicate that $c_k$ is positively causal ($ACE > 0$) for pixels at spatial locations which correspond to the $k^{th}$ digit. This agrees with the causal structure (by construction of VAE, $c_k$ causes the $k^{th}$ digit image). Figs \ref{decoder images}b, c and d correspond respectively to ACE of $z_0, z_6, \& z_2$  with intervened values ($\alpha$) increased from -3.0 to 3.0 ($z_0 \sim \N(0,1)$, so $3\sigma$ deviations) and $c_8=1$. The latents $z_0$ and $z_6$ seem to control the rotation and scaling of the digit 8 respectively. All other $z_k$'s behave similar to the plots for $z_2$, with no discernable causal effect on the decoded image. These observations are consistent with visual inspection on the decoded images after intervening on the latent space. More results with similar trends are reported in Appendix \ref{app_subsec_addl_results_vce}. 

\vspace{-4pt}
\section{Conclusions}
\label{sec_conclusions}
\vspace{-4pt}
This work presented a new causal perspective to neural network attribution. The presented approach views a neural network as an SCM, and introduces an appropriate definition, as well as a mechanism to compute, Average Causal Effect (ACE) effectively in neural networks. The work also presents a strategy to efficiently compute ACE for high-dimensional data, as well as extensions of the methodology to RNNs. The experiments on synthetic and real-world data show significant promise of the methodology to elicit causal effect of input on output data in a neural network. Future work will include extending to other neural network architectures (such as ConvNets) as well as studying the impact of other baselines on the proposed method's performance. Importantly, we believe this work can encourage viewing a neural network model from a causal lens, and answering further causal questions such as: which counterfactual questions might be asked and answered in a neural network causal model, can a causal chain exist in a neural network, are predictions made by neural networks causal, and so on.

\vspace{-6pt}

\section*{Acknowledgements}
We are grateful to the Ministry of Human Resource Development, India; Department of Science and Technology, India; as well as Honeywell India for the financial support of this project through the UAY program. We thank the anonymous reviewers for their valuable feedback that helped improve the presentation of this work.

\bibliography{example_paper}

\begin{thebibliography}{39}
\providecommand{\natexlab}[1]{#1}
\providecommand{\url}[1]{\texttt{#1}}
\expandafter\ifx\csname urlstyle\endcsname\relax
  \providecommand{\doi}[1]{doi: #1}\else
  \providecommand{\doi}{doi: \begingroup \urlstyle{rm}\Url}\fi

\bibitem[Abadi et~al.(2016)Abadi, Barham, Chen, Chen, Davis, Dean, Devin,
  Ghemawat, Irving, Isard, et~al.]{abadi2016tensorflow}
Abadi, M., Barham, P., Chen, J., Chen, Z., Davis, A., Dean, J., Devin, M.,
  Ghemawat, S., Irving, G., Isard, M., et~al.
\newblock Tensorflow: A system for large-scale machine learning.
\newblock In \emph{OSDI}, volume~16, pp.\  265--283, 2016.

\bibitem[Alvarez-Melis \& Jaakkola(2017)Alvarez-Melis and
  Jaakkola]{alvarez2017causal}
Alvarez-Melis, D. and Jaakkola, T.~S.
\newblock A causal framework for explaining the predictions of black-box
  sequence-to-sequence models.
\newblock \emph{arXiv preprint arXiv:1707.01943}, 2017.

\bibitem[Alvarez-Melis \& Jaakkola(2018)Alvarez-Melis and
  Jaakkola]{alvarez2018towards}
Alvarez-Melis, D. and Jaakkola, T.~S.
\newblock Towards robust interpretability with self-explaining neural networks.
\newblock \emph{arXiv preprint arXiv:1806.07538}, 2018.

\bibitem[Bach et~al.(2015)Bach, Binder, Montavon, Klauschen, M{\"u}ller, and
  Samek]{bach2015pixel}
Bach, S., Binder, A., Montavon, G., Klauschen, F., M{\"u}ller, K.-R., and
  Samek, W.
\newblock On pixel-wise explanations for non-linear classifier decisions by
  layer-wise relevance propagation.
\newblock \emph{PloS one}, 10\penalty0 (7):\penalty0 e0130140, 2015.

\bibitem[Bongers et~al.(2016)Bongers, Peters, Sch{\"o}lkopf, and
  Mooij]{bongers2016structural}
Bongers, S., Peters, J., Sch{\"o}lkopf, B., and Mooij, J.~M.
\newblock Structural causal models: Cycles, marginalizations, exogenous
  reparametrizations and reductions.
\newblock \emph{arXiv preprint arXiv:1611.06221}, 2016.

\bibitem[Christopher(2016)]{christopher2016pattern}
Christopher, M.~B.
\newblock \emph{PATTERN RECOGNITION AND MACHINE LEARNING.}
\newblock Springer-Verlag New York, 2016.

\bibitem[Claeskens et~al.(2008)Claeskens, Hjort, et~al.]{claeskens2008model}
Claeskens, G., Hjort, N.~L., et~al.
\newblock Model selection and model averaging.
\newblock \emph{Cambridge Books}, 2008.

\bibitem[Daniusis et~al.(2010)Daniusis, Janzing, Mooij, Zscheischler, Steudel,
  Zhang, and Schölkopf]{Daniusis2010}
Daniusis, P., Janzing, D., Mooij, J., Zscheischler, J., Steudel, B., Zhang, K.,
  and Schölkopf, B.
\newblock Inferring deterministic causal relations.
\newblock pp.\  143--150, 01 2010.

\bibitem[Deng et~al.(2014)Deng, Yu, et~al.]{deng2014deep}
Deng, L., Yu, D., et~al.
\newblock Deep learning: methods and applications.
\newblock \emph{Foundations and Trends{\textregistered} in Signal Processing},
  7\penalty0 (3--4):\penalty0 197--387, 2014.

\bibitem[Dheeru \& Karra~Taniskidou(2017)Dheeru and Karra~Taniskidou]{Dua:2017}
Dheeru, D. and Karra~Taniskidou, E.
\newblock {UCI} machine learning repository, 2017.
\newblock URL \url{http://archive.ics.uci.edu/ml}.

\bibitem[Eberhardt(2007)]{eberhardt2007causation}
Eberhardt, F.
\newblock Causation and intervention.
\newblock \emph{Unpublished doctoral dissertation, Carnegie Mellon University},
  2007.

\bibitem[Frosst \& Hinton(2017)Frosst and Hinton]{frosst2017distilling}
Frosst, N. and Hinton, G.
\newblock Distilling a neural network into a soft decision tree.
\newblock \emph{arXiv preprint arXiv:1711.09784}, 2017.

\bibitem[Geiger et~al.(1990)Geiger, Verma, and Pearl]{geiger1990identifying}
Geiger, D., Verma, T., and Pearl, J.
\newblock Identifying independence in bayesian networks.
\newblock \emph{Networks}, 20\penalty0 (5):\penalty0 507--534, 1990.

\bibitem[Gilmer et~al.(2017)Gilmer, Schoenholz, Riley, Vinyals, and
  Dahl]{gilmer2017neural}
Gilmer, J., Schoenholz, S.~S., Riley, P.~F., Vinyals, O., and Dahl, G.~E.
\newblock Neural message passing for quantum chemistry.
\newblock \emph{arXiv preprint arXiv:1704.01212}, 2017.

\bibitem[Higgins et~al.(2016)Higgins, Matthey, Pal, Burgess, Glorot, Botvinick,
  Mohamed, and Lerchner]{higgins2016beta}
Higgins, I., Matthey, L., Pal, A., Burgess, C., Glorot, X., Botvinick, M.,
  Mohamed, S., and Lerchner, A.
\newblock beta-vae: Learning basic visual concepts with a constrained
  variational framework.
\newblock 2016.

\bibitem[Hochreiter \& Schmidhuber(1997)Hochreiter and
  Schmidhuber]{hochreiter1997long}
Hochreiter, S. and Schmidhuber, J.
\newblock Long short-term memory.
\newblock \emph{Neural computation}, 9\penalty0 (8):\penalty0 1735--1780, 1997.

\bibitem[Hoyer et~al.(2009)Hoyer, Janzing, Mooij, Peters, and
  Sch{\"o}lkopf]{hoyer2009nonlinear}
Hoyer, P.~O., Janzing, D., Mooij, J.~M., Peters, J., and Sch{\"o}lkopf, B.
\newblock Nonlinear causal discovery with additive noise models.
\newblock In \emph{Advances in neural information processing systems}, pp.\
  689--696, 2009.

\bibitem[Hyttinen et~al.(2013)Hyttinen, Eberhardt, and
  Hoyer]{hyttinen2013experiment}
Hyttinen, A., Eberhardt, F., and Hoyer, P.~O.
\newblock Experiment selection for causal discovery.
\newblock \emph{The Journal of Machine Learning Research}, 14\penalty0
  (1):\penalty0 3041--3071, 2013.

\bibitem[Kiiveri et~al.(1984)Kiiveri, Speed, and Carlin]{kiiveri1984recursive}
Kiiveri, H., Speed, T.~P., and Carlin, J.~B.
\newblock Recursive causal models.
\newblock \emph{Journal of the australian Mathematical Society}, 36\penalty0
  (1):\penalty0 30--52, 1984.

\bibitem[Kindermans et~al.(2017)Kindermans, Hooker, Adebayo, Alber, Sch{\"u}tt,
  D{\"a}hne, Erhan, and Kim]{kindermans2017reliability}
Kindermans, P.-J., Hooker, S., Adebayo, J., Alber, M., Sch{\"u}tt, K.~T.,
  D{\"a}hne, S., Erhan, D., and Kim, B.
\newblock The (un) reliability of saliency methods.
\newblock \emph{arXiv preprint arXiv:1711.00867}, 2017.

\bibitem[Kingma et~al.(2014)Kingma, Mohamed, Rezende, and
  Welling]{kingma2014semi}
Kingma, D.~P., Mohamed, S., Rezende, D.~J., and Welling, M.
\newblock Semi-supervised learning with deep generative models.
\newblock In \emph{Advances in neural information processing systems}, pp.\
  3581--3589, 2014.

\bibitem[Kocaoglu et~al.(2017)Kocaoglu, Snyder, Dimakis, and
  Vishwanath]{kocaoglu2017causalgan}
Kocaoglu, M., Snyder, C., Dimakis, A.~G., and Vishwanath, S.
\newblock Causalgan: Learning causal implicit generative models with
  adversarial training.
\newblock \emph{arXiv preprint arXiv:1709.02023}, 2017.

\bibitem[Letham et~al.(2015)Letham, Rudin, McCormick, Madigan,
  et~al.]{letham2015interpretable}
Letham, B., Rudin, C., McCormick, T.~H., Madigan, D., et~al.
\newblock Interpretable classifiers using rules and bayesian analysis: Building
  a better stroke prediction model.
\newblock \emph{The Annals of Applied Statistics}, 9\penalty0 (3):\penalty0
  1350--1371, 2015.

\bibitem[Li et~al.(2018)Li, Liu, Chen, and Rudin]{li2018deep}
Li, O., Liu, H., Chen, C., and Rudin, C.
\newblock Deep learning for case-based reasoning through prototypes: A neural
  network that explains its predictions.
\newblock In \emph{Thirty-Second AAAI Conference on Artificial Intelligence},
  2018.

\bibitem[Montavon et~al.(2017)Montavon, Lapuschkin, Binder, Samek, and
  M{\"u}ller]{montavon2017explaining}
Montavon, G., Lapuschkin, S., Binder, A., Samek, W., and M{\"u}ller, K.-R.
\newblock Explaining nonlinear classification decisions with deep taylor
  decomposition.
\newblock \emph{Pattern Recognition}, 65:\penalty0 211--222, 2017.

\bibitem[Pearl(2009)]{pearl2009causality}
Pearl, J.
\newblock \emph{Causality}.
\newblock Cambridge university press, 2009.

\bibitem[Pearl(2012)]{pearl2012calculus}
Pearl, J.
\newblock The do-calculus revisited.
\newblock \emph{arXiv preprint arXiv:1210.4852}, 2012.

\bibitem[Peters et~al.(2017)Peters, Janzing, and
  Sch{\"o}lkopf]{peters2017elements}
Peters, J., Janzing, D., and Sch{\"o}lkopf, B.
\newblock \emph{Elements of causal inference: foundations and learning
  algorithms}.
\newblock MIT press, 2017.

\bibitem[Ribeiro et~al.(2016)Ribeiro, Singh, and Guestrin]{ribeiro2016should}
Ribeiro, M.~T., Singh, S., and Guestrin, C.
\newblock Why should i trust you?: Explaining the predictions of any
  classifier.
\newblock In \emph{Proceedings of the 22nd ACM SIGKDD International Conference
  on Knowledge Discovery and Data Mining}, pp.\  1135--1144. ACM, 2016.

\bibitem[Sadowski et~al.(2014)Sadowski, Whiteson, and
  Baldi]{sadowski2014searching}
Sadowski, P.~J., Whiteson, D., and Baldi, P.
\newblock Searching for higgs boson decay modes with deep learning.
\newblock In \emph{Advances in Neural Information Processing Systems}, pp.\
  2393--2401, 2014.

\bibitem[Selvaraju et~al.(2016)Selvaraju, Das, Vedantam, Cogswell, Parikh, and
  Batra]{selvaraju2016grad}
Selvaraju, R.~R., Das, A., Vedantam, R., Cogswell, M., Parikh, D., and Batra,
  D.
\newblock Grad-cam: Why did you say that?
\newblock \emph{arXiv preprint arXiv:1611.07450}, 2016.

\bibitem[Shrikumar et~al.(2017)Shrikumar, Greenside, and
  Kundaje]{shrikumar2017learning}
Shrikumar, A., Greenside, P., and Kundaje, A.
\newblock Learning important features through propagating activation
  differences.
\newblock \emph{arXiv preprint arXiv:1704.02685}, 2017.

\bibitem[Simonyan et~al.(2013)Simonyan, Vedaldi, and
  Zisserman]{simonyan2013deep}
Simonyan, K., Vedaldi, A., and Zisserman, A.
\newblock Deep inside convolutional networks: Visualising image classification
  models and saliency maps.
\newblock \emph{arXiv preprint arXiv:1312.6034}, 2013.

\bibitem[Smilkov et~al.(2017)Smilkov, Thorat, Kim, Vi{\'e}gas, and
  Wattenberg]{smilkov2017smoothgrad}
Smilkov, D., Thorat, N., Kim, B., Vi{\'e}gas, F., and Wattenberg, M.
\newblock Smoothgrad: removing noise by adding noise.
\newblock \emph{arXiv preprint arXiv:1706.03825}, 2017.

\bibitem[Sundararajan et~al.(2017)Sundararajan, Taly, and
  Yan]{sundararajan2017axiomatic}
Sundararajan, M., Taly, A., and Yan, Q.
\newblock Axiomatic attribution for deep networks.
\newblock \emph{arXiv preprint arXiv:1703.01365}, 2017.

\bibitem[Team(2017)]{team2017pytorch}
Team, P.~C.
\newblock Pytorch: Tensors and dynamic neural networks in python with strong
  gpu acceleration, 2017.

\bibitem[Yosinski et~al.(2015)Yosinski, Clune, Nguyen, Fuchs, and
  Lipson]{yosinski2015understanding}
Yosinski, J., Clune, J., Nguyen, A., Fuchs, T., and Lipson, H.
\newblock Understanding neural networks through deep visualization.
\newblock \emph{arXiv preprint arXiv:1506.06579}, 2015.

\bibitem[Zeiler \& Fergus(2014)Zeiler and Fergus]{zeiler2014visualizing}
Zeiler, M.~D. and Fergus, R.
\newblock Visualizing and understanding convolutional networks.
\newblock In \emph{European conference on computer vision}, pp.\  818--833.
  Springer, 2014.

\bibitem[Zhou \& Troyanskaya(2015)Zhou and Troyanskaya]{zhou2015predicting}
Zhou, J. and Troyanskaya, O.~G.
\newblock Predicting effects of noncoding variants with deep learning--based
  sequence model.
\newblock \emph{Nature methods}, 12\penalty0 (10):\penalty0 931, 2015.

\end{thebibliography}
\bibliographystyle{icml2019}

\clearpage
\appendix




\section{Appendix}
\subsection{Causality Preliminaries}
\label{Basics of causality}
In this section, we review some of the basic definitions in causality that may help understand this work.\\

Structural Causal Models (SCMs) \cite{pearl2009causality} provide a rigorous definition of cause-effect relations between different random variables. Exogenous variables (noise) are the only source of stochasticity in an SCM, with the endogenous variables (observables) deterministically fixed via functions over the exogenous and other endogenous variables.
\theoremstyle{definition}
\begin{definition}({Structural Causal Models}). A Structural Causal Model is a 4-tuple $(X, U, f, P_u))$ where, (i) $X$ is a finite set of endogenous variables, usually the observable random variables in the system; (ii) $U$ is a finite set of exogenous variables, usually treated as unobserved or noise variables; (iii) $f$ is a set of functions $[f_1, f_2,....f_n]$, where n refers to the cardinality of the set $X$. These functions define causal mechanisms, such that $\forall  x_i \in X, x_i = f_i(Par,u_i)$. The set $Par$ is a subset of $X - \{x_i\}$ and $u_i \in U$. We do not consider feedback causal models here; (iv) $P_u$ defines a probability distribution over $U$. It is not necessary for every node in an SCM to have a unique/shared noise. Deterministic causal systems have been considered in literature \cite{Daniusis2010}.

\label{definition:SCM}
\end{definition}

An SCM $M(X, U, f, P_u)$ can be trivially represented by a directed graphical model $G=(V,E)$, where the vertices $V$ represent the endogenous variables $X$ (each vertex $v_i$ corresponds to an observable $x_i$). We will use random variables and vertices interchangeably henceforth. The edges $E$ denote the causal mechanisms $f$. Concretely, if $x_i = f_i(Par,u_i)$ then $\forall x_j \in Par$, there exists a directed edge from the vertex $v_j$ corresponding to $x_j$ to the vertex $v_i$ corresponding to $x_i$. The vertex $v_j$ is called the \textit{parent} vertex while the vertex $v_i$ is referred to as the \textit{child} vertex. Such a graph is called a \textbf{causal Bayesian network}. The distribution of every vertex in a causal Bayesian network depends only upon its parent vertices (local Markov property) \cite{kiiveri1984recursive}. 

A \textit{path} is defined as a sequence of unique vertices $v_o, v_1, v_2, ..., v_n$ with edges  between each consecutive vertex $v_i$ and $v_{i+1}$. A {collider} is defined with respect to a path as a vertex  $v_i$ which has a $\rightarrow v_i \leftarrow$ structure. (The direction of the arrows imply the direction of the edges along the path.) \textit{d-separation} is a well-studied property of graphical models \cite{pearl2009causality, geiger1990identifying} that is often used to decipher conditional independences between random variables that admit a probability distribution faithful to the graphical model. 

\begin{propositions}\cite{pearl2009causality}
Two random variables $a$ and $b$ are said to  be conditionally independent given a set of random variables $Z$ if they are \textit{d-separated} in the corresponding graphical model $G$.
\label{prop: conditional_independence}
\end{propositions}

\theoremstyle{definition}
\begin{definition}({d-separation}).
Two vertices $v_a$ and $v_b$ are said to be \textit{d-separated} if all paths connecting the two vertices are ``blocked'' by a set of random variables $Z$. 
\label{def: d-separation}
\end{definition}

A path is said to be ``blocked'' if either (i) there exists a \textit{collider} that is not in $Anc(Z)$, or, (ii) there exists a \textit{non-collider} $v \in Z$ along the path.  $Anc(Z)$ is the set of all vertices which exhibit a \textit{directed path} to any vertex $v \in Z$. A \textit{directed path} from vertex $v_i$ to $v_j$ is a \textit{path} such that there is no incoming edge to $v_i$ and no outgoing edge from $v_j$.

The $do(.)$ operator (Definition \ref{def: ACE})\cite{pearl2009causality, pearl2012calculus} is used to identify causal effects from a given SCM or causal Bayesian network. Although similar in appearance to the conditional expectation $\E(y|x = 1)$, $\E(y|do(x) = 1)$ refers to the expectation of the random variable $y$ taken over its interventional distribution $P(y|do(x) = 1)$. 

\theoremstyle{definition}
\begin{definition}({Average Causal Effect}).
The Average Causal Effect (ACE) of a binary random variable $x$ on another random variable $y$ is commonly defined as $\E(y|do(x = 1)) - \E(y|do(x = 0))$.
\end{definition}

Formally, a causal Bayesian network $G = (V,E)$ induces a joint distribution over its vertices $P_{V} = \prod_{v_i \in V}P(v_i|parents(v_i)$. Performing interventions on random variables $X_i$ are analogous to surgically removing incoming edges to their corresponding vertices $V_{X_i}$ in the network $G$. This is because the value of the random variables $X_i$ now depend on the nature of the intervention caused by the ``external doer'' and not the inherent causal structure of the system. The interventional joint distribution over the vertices of $G$ would be $P_{(V | do(V_{X_i}))} = \prod_{v_i \in V - V_{X_i}}P(v_i|parents(v_i)))$. 
Notice that in $P_{(V | do(V_{X_i}))}$, the factorization of the interventional joint distribution ignores the intervened random variables $X_i$. In an SCM $M(X, U, f, P_u)$, performing a $do(x = x')$ operation is the same as an intervened SCM $M^i(X, U, f^i, P_u)$, where the causal mechanism $f_x$ for variable x, is replaced by the constant function $x'$. $f^i$ is obtained from the set $f$ by replacing all the instances of random variable $x$ in the arguments of the causal functions by $x'$.

\subsection{More on Prior Work}
\label{app_subsec_prior_work}
Existing methods for attribution can broadly be categorized into gradient-based methods and local regression-based methods. 

As stated in Sections \ref{introduction} and \ref{Section: previous_work} (main paper), in the former approach, gradients of a function are not ideal indicators of an input feature's influence on the output. Partial derivatives of a continuous function $f:\R^n \rightarrow \R$ are also functions $g_i:\R^n \rightarrow \R$ over the same domain $\R^n$ (the subscript $i$ denotes the partial derivative with respect to the $i^{th}$ input feature). The attribution value of the $i^{th}$ feature which is derived from $g_i$ would in turn be biased by the values of other input features. For instance, consider a simple function $f:\R^2 \rightarrow \R$, $f(x_1,x_2) = x_1x_2$. The respective partial derivatives are  $g_1 = x_2$ and $g_2 = x_1$. Consider a points $a = [5,1000]$. $g_1(a) = 1000$ and $g_2(a) = 5$. This implies that for output $f(a) = 5000$, $x_1$ had a stronger influence than $x_2$. But in reality $x_2$ has a stronger contribution towards $f(a)$ than $x_1$. Gradients are thus viable candidates for the question ``How much would perturbing a particular input affect the output?'', but not for determining which input influenced a particular output neuron.

Besides, perturbations and gradients can be viewed as capturing the Individual Causal Effect (ICE) of input neuron $x_i$ with values $\alpha$ on output $y$. 
\begin{equation}
ICE^y_{do(x_i = \alpha)} = \E[y|do(x_i = \alpha), x_{j \neq i} = data] - baseline
\label{ICE}
\end{equation}
In Equation \ref{ICE}, $x_{j \neq i} = data$ denotes conditioning the input neurons other than $x_i$ to the input training instance values. The Expectation operator for $y$ is over the unobservable noise which is equal to the learned neural function $f(.)$ itself, i.e., 
$ICE^y_{do(x_i = \alpha)} = f(x_1, x_2, ..., \alpha .., x_n) - baseline$, where the baseline is $f(x_1, x_2, ..., \alpha - \epsilon, .., x_n)$ for some $\epsilon \in {\rm I\!R}$. Evidently, inter-feature interactions can conceal the real importance of input feature $x_i$ in this computation, when only the ICE is analyzed.

The latter approach of ``interpretable" regression is highly prone to artifacts as regression primarily maps correlations rather than causation. Regression of an output variable $y$ (the neural network output) on a set of input features is akin to calculating $\E[y|x_1, x_2, ..., x_k]$, given $k$ input features. However, true causal effects of $x_i$ on $y$ are discerned via $\E[y| do(x_i)]$, as in \cite{pearl2009causality}. The only way regressing on a particular input feature would give $\E[y| do(x_i)]$ is if all the backdoor variables are controlled and a weighted average according to the distribution of these backdoor variables is taken \cite{pearl2009causality}. Thus, causal statements made from regressing on all input variables (say, the weights of a linear approximator to a deep network) would be far from the true picture.

\subsection{Proofs}
\subsubsection{Proof of Proposition \ref{prop: NN as SCM}}
\label{appendix: proposition 2}
\begin{proof}
In a feedforward neural network, each layer neurons can be written as functions of neurons in its previous layer, i.e. $\forall i \in l: \forall l_{i_j} \in l_{i}: l_{i_j} = f_{i_j}(l_{i - 1})$. The input layer $l_1$ can be assumed to be functions of independent noise variables $U$ such that $l_{1_i} = f_{1_i}(u_i) $ $\forall l_{1_i} \in l_1$ and $u_i \in U$. This structure in the random variables, neurons in the network, can be equivalently expressed by a SCM $M([l_1,  l_2, ...., l_n], U, [f_1, f_2, ... f_n], P_u)$.
\end{proof}

\subsubsection{Proof of Corollary \ref{corr: recursive_subs}}
\label{appendix: corollary 2}
\begin{proof}
All notations are consistent with their definitions in Proposition \ref{prop: NN as SCM}. Starting with each neuron $l_{n_i}$ in the output layer $l_n$, the corresponding causal function $f_{n_i}(l_{n - 1})$ can be substituted as $f_{n_i}(f_{{n - 1}_1}(l_{n - 2}), f_{{n - 1}_2}(l_{n - 2}), f_{{n - 1}_3}(l_{n - 2}), ... f_{{n - 1}_{|l_{n-1|}}}(l_{n - 2}))$. This can also be written as $l_{n_i} = f'_{n_i}(l_{n - 2})$. $f_{i_j}$ refers to the causal function of neuron $j$ in layer $i$. Similarly, $l_{i_j}$ refers to neuron $j$ in layer $i$. Proceeding recursively layer by layer, we obtain modified functions such that, $\forall l_{n_i} \in layer$ $l_n: l_{n_i} = f'_{n_i}(l_1)$. The causal mechanisms set $f'$ of the reduced SCM M' would be $\{f'_{n_i}| l_{n_i} \in l_n\} \cup $\{$l_{1_i} = f_{1_i}(u_i)$ $| l_{1_i} \in l_1$ and $u_i \in U$$\}$  
\end{proof}

\subsubsection{Proof of Proposition \ref{prop: d-sep feedforward}}
\label{appendix: proposition 3}
\begin{proof}
Let $M^c$ be the causally sufficient SCM for a given SCM $M'$. Let $G^c = (V,E)$ be the corresponding causal bayesian network. Presence of dependency between input features in neural network $N$ implies the existence of common exogenous parent vertices in the graph $G^c$. All the paths from one input neuron to another in graph $G^c$ either passes through an exogenous variable or a vertex corresponding to an output neuron. The output neurons are colliders and the intervention on $v_i$, surgically removes all incoming edges to $v_i$ (refer to Section \ref{Basics of causality}). As all the paths from $v_i$ to every other input neuron $v_j$ are ``blocked'', from Definition \ref{def: d-separation}, the intervened input neuron is d-seperated from all other input neurons.    
\end{proof}

\subsubsection{Proof of Proposition \ref{proposition: get tau}}
\label{proof proposition 4}
\begin{proof}
Let $p_{y^t}$ be a probability density over the output variables $y^t$ at time $t$.
Now, from Corollary \ref{corr: recursive_subs} and Section \ref{Section: philosophy}
\begin{equation}
y^t = f(x^1, x^2, ..., x^{t - 1}). 
\end{equation}
$f(.)$ is a recurrent function (the neural network). 
\end{proof}
In the reduced SCM $M'$ for the recurrent function $f(.)$, if the values of all other input neurons at different timesteps are controlled (fixed), $y^t$ transforms according to $f(x^{t-k})$. Let's assume $y^t$ depends on $x^{t-k}$ via a one-to-one mapping. Note, if there exists a one-to-one mapping between $x^{t - k}$ and $y^t$, then the conditional entropy $H(y^t|x^{t - k})$ would be $0$, thus maximizing the mutual information between the two random variables. So, we limit the lookback to only those timesteps that register a one-to-one mapping with $y^t$.

The probability of $y^t$ in an infinitesimal volume $dy^t$ is given by, 
\begin{equation}
P(y^t) = p(y^t)dy^t
\end{equation}
By change of variables 
\begin{equation}
P(y^t) = p(y^t(x^{t - k}))|det(\nabla_{x^{t - k}} y^t)|dx^{t - k}
\label{determinat proportion}
\end{equation}

Now, $dy^t$ and $dx^{t - k}$ are volumes and hence are positive constants. $y^t$ exists in the training data and hence $P(y^t) > 0$. Similarly, $p(y^t(x^{t - k})) \neq 0$. Thus, if $P(y^t)$ evaluated using Equation \ref{determinat proportion} is zero, there is a contradiction. Hence, the assumption that $y^t$ depends on $x^{t-k}$ via a one-to-one mapping is incorrect. $\tau_x = \max_k(|det(\nabla_{x^{t - k}}y^t)| > 0)$. would be optimal for a particular input sequence $x$ and output $y^t$.
$\E_x[\max_k(|det(\nabla_{x^{t - k}}y^t)| > 0)]$ is taken as the $\tau$ for the entire dataset, to prevent re-computation for every new input sequence.

\subsection{Algorithms/Pseudocode}
\subsubsection{Algorithm for Phase I in Feedforward Networks}
\label{Algorithm for Phase 1 in case of Feedforward networks}
\begin{algorithm}[tb]
\caption{Calculate interventional expectation for feedforward networks}
   \label{Calculate interventional expectation}
\begin{algorithmic}
   \STATE {\bfseries Result:} $\E(y|do(x_i))$
   \STATE {\bfseries Input:} output neuron $y$, intervened input neuron $x_i$, input value constraints [$low^i, high^i$], number of interventions $num$, means $\mu$, covariance matrix $Cov$, neural network function $f()$
   \STATE {\bfseries Initialize:} $Cov[x_i][:] := 0$; $Cov[:][x_i] := 0$; $interventional\_expectation := []$; $\alpha = low^i$ 

   \WHILE{$\alpha \leq high^i$}
   \STATE $\mu[i]$ := $\alpha$ 
   \STATE $interventional\_expectation$.append($f(\mu)$+ $\frac{1}{2}$trace(matmul($\nabla^2f(\mu),Cov$)))\;
   \STATE $\alpha := \alpha + \frac{high^i - low^i}{num}$ \;
   \ENDWHILE

\end{algorithmic}
\end{algorithm}

Algorithm \ref{Calculate interventional expectation} outputs an array of size $num$ with interventional expectations of an output neuron $y$ given different interventions ($do(.)$) on $x_i$. The user input parameter $num$ decides how many evenly spaced $\alpha$ values are desired. The accuracy of the learned polynomial functions in Phase II depends on the size of $num$. 

Consider $n$ training points, and $k$ input neurons in a feedforward network. Usually, $n \gg k$  to avoid memorization by the network. Computations are performed on-the-fly via a single pass through the computational graph in frameworks such as  Tensorflow \cite{abadi2016tensorflow} and PyTorch\cite{team2017pytorch}. If one single pass over the computational graph is considered $1$ unit of computation, the computational complexity of Phase I (Algorithm \ref{Calculate interventional expectation}) would be O($k \times num$). Compare this to the computational complexity of O($n \times num$) for calculating the interventional expectations naively. For every perturbation $\alpha$ of neuron $x_i$, we would require atleast $n$ forward passes on the network to estimate $\E(y|do(x_i = \alpha))$.


\subsubsection{Algorithm for Phase I in Recurrent networks}
\label{Algorithm for Phase 1 in case of Recurrent networks}
See Algorithm \ref{algo: phase I recurrent}. The input training data is arranged in a tensor of size $num\_samples \times num\_time \times num\_features$.
\begin{algorithm}[tb]
\caption{Calculate interventional expectation for recurrent networks}
   \label{Calculate interventional expectation recurrent}
\begin{algorithmic}
   \STATE {\bfseries Result:} $\E(y^t|do({x_i^{\hat{t}}}))$
   \STATE {\bfseries Input:} output neuron $y^t$, intervened input neuron $x_i^{\hat{t}}$ at time $\hat{t}$, input value constraints [$low_i^{\hat{t}}, high_i^{\hat{t}}$], number of interventions $num$, training input data $Data$, recurrent function $f()$
   \STATE {\bfseries Initialize:} $\alpha=low_i^{\hat{t}}$; $interventional\_expectation := []$; 
   \WHILE{$\alpha \leq high_i^{\hat{t}}$}
   \STATE $data\_iterator$ := 0
   \STATE $inputdata$ := $Data[:,:t+1,:]$ //past is independent of the present timestep $t$
   \STATE $inputdata[:,t,i]$ := $\alpha$ //setting the value of the intervened variable\;
   \WHILE{$data\_iterator < Data.size()$}
   \STATE $next\_timestep\_input$ := $f(input\_data)$
   \STATE $inputdata$.append($next\_timestep\_input$)
   \STATE $data\_iterator$ += 1
   \ENDWHILE
   \STATE $\mu$ := Mean($inputdata$) //Calculate mean of each input neuron\;
   \STATE $Cov$ := Covariance($inputdata$)
   \STATE $tempvar$ := $f(\mu)$
   \STATE $hess$ := $\nabla^2f(\mu)$
   \STATE $interventional\_expectation$.append($tempvar$ + $\frac{1}{2}$trace(matmul($hess$,$Cov$)))\;  
   \STATE $\alpha := \alpha + \frac{high^i - low^i}{num}$
   \ENDWHILE

\end{algorithmic}
\label{algo: phase I recurrent}
\end{algorithm}

\subsubsection{Algorithm for Phase II}
\label{phase II algorithm}
See Algorithm \ref{algo: phase II}.
\begin{algorithm}[tb]
\caption{Learning causal regressors}
   \label{learn polynomials}
\begin{algorithmic}
   \STATE {\bfseries Result:} $baseline, predictive\_mean, predictive\_variance$
   \STATE {\bfseries Input:} interventional expectation for different interventions $\E(y|do(x_i))$, input value constraints [$low^i, high^i$]
   \STATE {\bfseries Initialize:} $\alpha=low^i$;
   \STATE order := Bayesian\_Model\_Selection($\E(y|do(x_i))$) \;
   \STATE predictive\_mean, predictive\_variance := Bayesian\_linear\_regression($\E(y|do(x_i))$, order)\;
   \STATE baseline := Integrate(predictive\_mean, $low^i, high^i$)
\end{algorithmic}
\label{algo: phase II}
\end{algorithm}

\subsection{Scaling to Large Data}
\label{section: scalability}
\vspace{-4pt}
In this section we follow the same notations as defined in Section \ref{Section: ACE_method} in the main text. Evaluating the interventional expectations using Eqn \ref{expected taylor expansion eq} involves calculating the Hessian. This is a costly operation. For a system with $k$ input features it takes about $O(k)$ backward passes along the computational graph. Several domains involve a large number of input features. Such a large $k$ regime would render Equation \ref{expected taylor expansion eq}  inefficient. Note however that we never explicitly require the Hessian, just the term $\sum_{i=1}^k\sum_{j=1}^k\nabla^2f'_y(\mu)_{ij}Cov(x_i, x_j|do(x_l = \alpha))$. In this section, we propose an efficient methodology to compute the interventional expectations for high-dimensional data.

We begin with computing $Cov(\mathbf{x},\mathbf{x}|do(x_l = \alpha))$, where $\mathbf{x}$ is the input vector. Consider the eigendecomposition of $Cov(\mathbf{x},\mathbf{x}|do(x_l = \alpha)) = \sum_{r=1}^k\lambda_r e_r e_r^T$, where $e_r$ is the $r^{th}$ eigenvector and $\lambda_r$ the corresponding eigenvalue. Let $v_r = \lambda^{1/2}e_r$. Performing a Taylor series expansion of $f'_y$ around $\mu$, we get: 
\vspace{-3pt}
\begin{equation*}
    \begin{aligned}
        f'_y(\mu + \epsilon v_r) &= f'_y(\mu) + \epsilon\nabla^Tf'_y(\mu)v_r + \frac{\epsilon^2}{2} v_r^T\nabla^2f'_y(\mu)v_r \\&+ O(\epsilon^3 v_r^3) 
    \end{aligned}
    \vspace{-4pt}
    \label{eq: taylor's expansion for eigenvectors_1}
\end{equation*}
\vspace{-3pt}
\begin{equation*}
    \begin{aligned}
        f'_y(\mu - \epsilon v_r) &= f'_y(\mu) - \epsilon\nabla^Tf'_y(\mu)v_r + \frac{\epsilon^2}{2} v_r^T\nabla^2f'_y(\mu)v_r \\&+ O(-\epsilon^3 v_r^3)  
    \end{aligned}
    \vspace{-4pt}
    \label{eq: taylor's expansion for eigenvectors_2}
\end{equation*}
Adding the equations: 
\vspace{-3pt}
\begin{equation*}
    \begin{aligned}
        f'_y(\mu - \epsilon v_r) +  f'_y(\mu + \epsilon v_r) - 2f'_y(\mu) &= \epsilon^2 v_r^T\nabla^2f'_y(\mu)v_r \\&+ O(\epsilon^4 v_r^4) \\ 
    \end{aligned}
    \vspace{-4pt}
\end{equation*}
\begin{equation*}
    \begin{aligned}
\frac{1}{\epsilon^2} \bigg(f'_y(\mu - \epsilon v_r) +  f'_y(\mu + \epsilon v_r) - 2f'_y(\mu)\bigg) &= v_r^T\nabla^2f'_y(\mu)v_r \\&+ O(\epsilon^2 v_r^4) \\
    \end{aligned}
    \vspace{-4pt}
\end{equation*}
\vspace{-4pt}
Rather:
\vspace{-4pt}
\begin{equation}
    \lim_{\epsilon \rightarrow 0}\frac{1}{\epsilon^2} \bigg(f'_y(\mu - \epsilon v_r) +  f'_y(\mu + \epsilon v_r) - 2f'_y(\mu)\bigg) = v_r^T\nabla^2f'_y(\mu)v_r
    \vspace{-4pt}
    \label{eq: directional hessian}
\end{equation}

Equation \ref{eq: directional hessian} calculates the second order directional derivative along $v_r$. Since $Cov(x_i, x_j|do(x_l = \alpha)) = \sum_{r=1}^k v_{ri}v_{rj}$ ($ri$ and $rj$ refer to the $i^{th}$ \& $j^{th}$ entry of $v_r$ respectively), $\sum_{r=1}^k v_r^T\nabla^2f'_y(\mu)v_r = \sum_{i=1}^k\sum_{j=1}^k\nabla^2f'_y(\mu)_{ij}Cov(x_i, x_j|do(x_l = \alpha))$. Thus, the second order term in Eqn \ref{expected taylor expansion eq} can be calculated by three forward passes on the computational graph with inputs $\mu, \mu + \epsilon V, \mu - \epsilon V$, where $V$ is the matrix with $v_r$s as columns and $\epsilon$ is taken to be very small ($10^{-6}$). Although eigendecomposition is also compute-intensive, the availability of efficient procedures allowed us to get results significantly faster than exact calculations (0.04s for the approximation v/s 3.04s per computation for experiments on MNIST dataset with a deep neural network of 4 hidden layers).

\begin{figure}
\centering
\includegraphics[scale=0.26]
{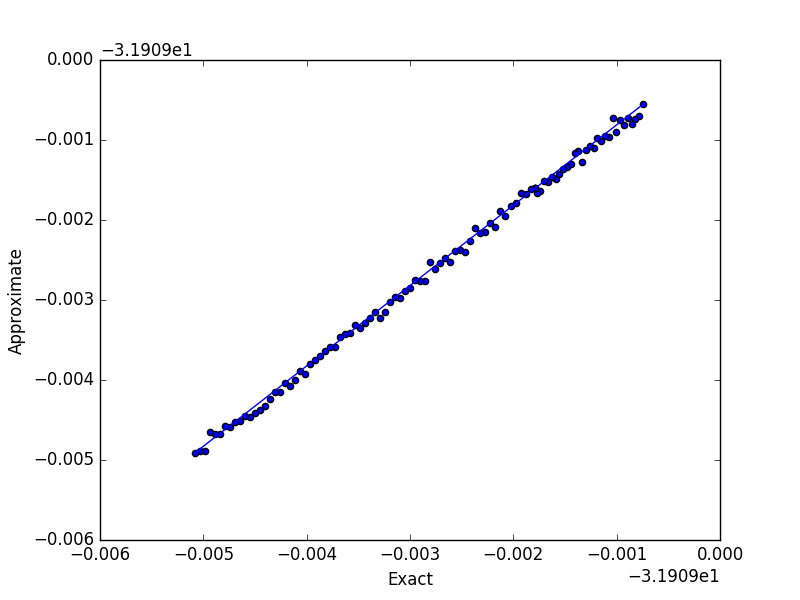}
  \caption{Quality of approximation via second order directional derivatives}
  \label{hess_approx_results}
\end{figure}

Figure \ref{hess_approx_results} shows results for the approximate second order term calculated v/s the exact second order term for different $\alpha$ values (Section \ref{section: scalability}). The function $f'_y$ is a neural network trained on MNIST images. Both the methods agree ``almost" perfectly with each other as indicated by the $y=x$ line.

In case of feedforward networks, from Corollary \ref{corr: independent nodes}, we know that $Cov(x_i, x_j|do(x_l = \alpha)) = Cov(x_i, x_j)$, i.e., the observational covariances. For recurrent networks, $Cov(x_i, x_j|do(x_l = \alpha))$ can be calculated after explicitly intervening on the system (Section \ref{subsec_causal_attrib_RNNs}). 

\subsection{More on Experiments and Results}

\subsubsection{Generation of Synthetic Dataset}
\label{generation of simulated dataset}
We used the following procedure for generating the synthetic dataset used for experiments (Section \ref{subsection: simulated_data}): 
\begin{itemize}
\item Sample individual sequences uniformly of length between $[T,T+5]$. We used $T = 10$. Let $x^t$ refer to the sequence value at length $t$. 
\item $\forall i; 2 < i \leq T$ Sample $x^i \sim \mathcal{N}(0,0.2)$. 
\item With probability $0.5$ either (a) sample $\forall i; 0   \leq i < 3$ $x^i \sim \mathcal{N}(1,0.2)$ and label such sequences class $1$  or (b) sample $\forall i; 0   \leq i < 3$ $x^i \sim \mathcal{N}(-1,0.2)$ and label such sequences class $0$. 
\end{itemize}

\subsubsection{Calculation of the interventional expectations in Section \ref{section: disentangled_latent_representations} }
\label{decoder_details}
From the generative model of the VAE we have access to $p(x_{ij}|z,c)$. Each pixel $p(x_{ij}|z,c)$ is modelled as a Bernoulli random variable with parameter $\theta_{ij}$, $z$ being the continous latents $[z_0, z_1, z_2, ..., z_9]$ and $c$ being the class-specific binary variables $[c_0, c_1, c_2, ..., c_9]$. Interventional expectations required for calculated ACEs are calculated via Equation \ref{interventional eq}.

\textit{For continuous latents:} 
    \begin{equation*}
    \begin{split}
        \E[x_{ij}|do(z_k= \alpha),do(c_l = 1)] = \\ \E_{z\backslash z_k}[\E_{x_{ij}}[x_{ij}|do(z_k= \alpha), do(c_l = 1), z]]. 
        \end{split}
    \end{equation*}
    From the generative model prior $p(z, c)$, we know that each $z_k$ is independently distributed according to $\N(0,1)$, so the intervention does not change the distribution of the other variables. However, the multinouli distribution over the $c'$s forces all the other $c_{i\neq l} = 0$. Thus, the above expression can be simply computed via Monte Carlo integration as follows:
    $\frac{1}{K}\Sigma_{{z\backslash z_k \sim N(0,I_{9})}}\theta_{ij}$, where $K$ samples are drawn.
    
\textit{For discrete latents:} there are two cases depending on the intervention value $\alpha$. \\
    Case 1:
   \begin{equation*}
        \E[x_{ij}|do(c_k = 1)] = 
        \E_{z}[\E_{x_{ij}}[x_{ij}|do(c_k = 1), z]]
   \end{equation*}
    As before, the multinoulli distribution over the $c$s restricts all the other $c_{i\neq k} = 0$. Thus, the above expression can be simply computed via Monte Carlo integration as follows:
    $\frac{1}{K}\Sigma_{{z \sim N(0,I_{10})}}\theta_{ij}$, where $K$ samples are drawn.\\
    Case 2:
    \begin{equation*}
        \E[x_{ij}|do(c_k = 0) = 
        \E_{z, c \backslash c_k}[\E_{x_{ij}}[x_{ij}|do(c_k = 0), c, z]]
   \end{equation*}
    Now, as $c_{k} = 0$, the distribution over all the other $c_{i \neq k} \sim Mult(1,\textit{U}\{0,9\}\backslash k)$. Thus, the above expression can be simply computed via Monte Carlo Integration as follows,
    $\frac{1}{K}\Sigma_{{z\sim N(0,I_{10})}{c \backslash c_k \sim Mult(1,\textit{U}\{0,9\}\backslash k)}}\theta_{ij}$, where $K$ samples are drawn.

\subsubsection{Additional Results: Visualizing Causal Effect}
\label{app_subsec_addl_results_vce}
\begin{figure*}
\centering
\includegraphics[scale=1]
{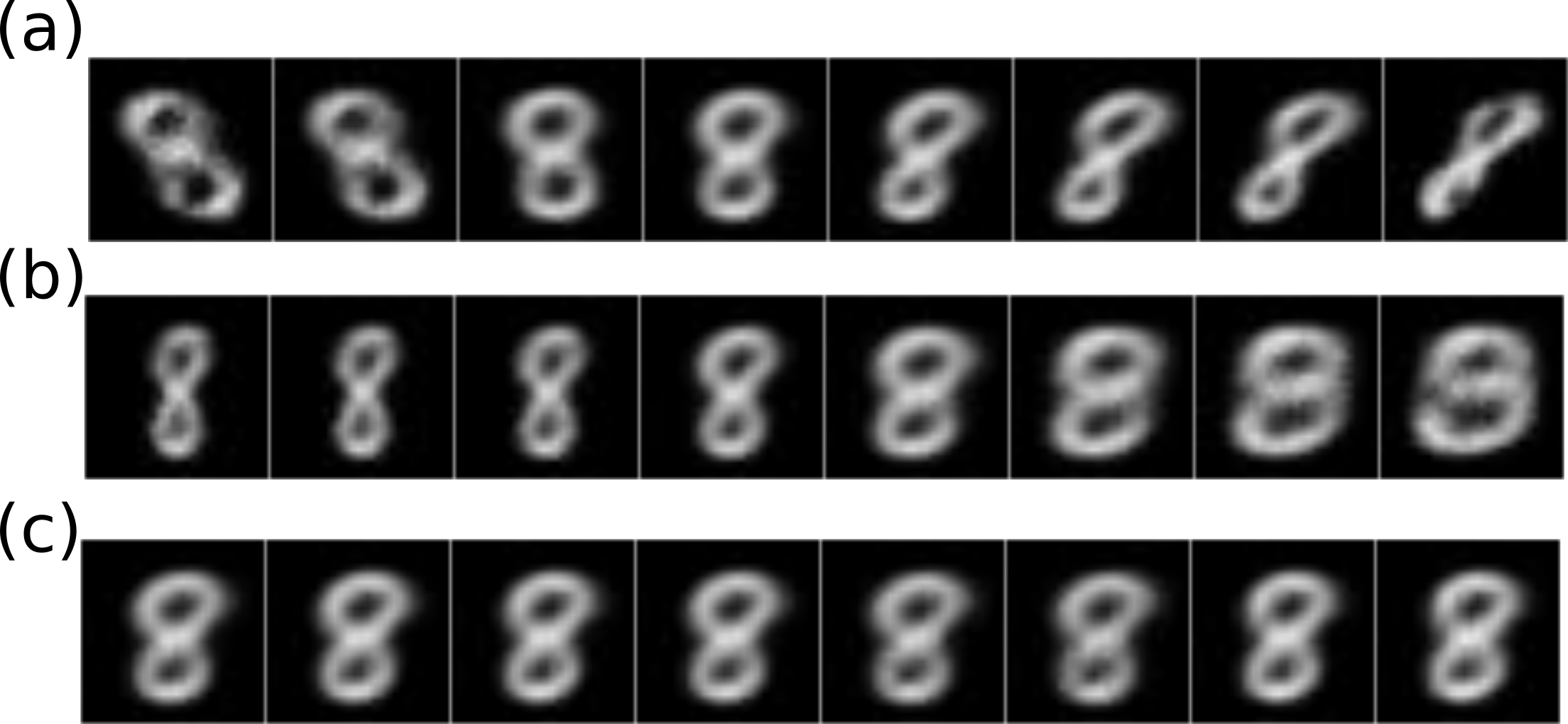}
  \caption{Decoded images generated by a random latent vector, with interventions between -3.0 to 3.0 on (a) $z_0$, (b) $z_6$, (c) $z_2$. The observed trends are consistent with the causal effects observed via causal attributions on the respective $z_k$s. $z_0$ captures rotation, $z_6$ captures scaling, and $z_2$ captures nothing discernable.}
  \label{decodede_8}
\end{figure*}

In continuation to results in Section \ref{section: disentangled_latent_representations}, we present additional results here. We fix the class part of the latent and sample a random vector z from $\N(0,1)$. Then we intervene on one of the dimensions of z and pass the latent through the decoder. We intervene with values in the range -3 to 3. This is repeated for every dimension. When the decoded images are sorted based on the value of intervention, we are able to see the effect of rotation in dimension $z_0$ and the effect of scaling in dimension $z_6$. Other dimensions show no effect, as shown in Figure \ref{decodede_8}.

\begin{figure*}
\centering
\includegraphics[scale=0.9]
{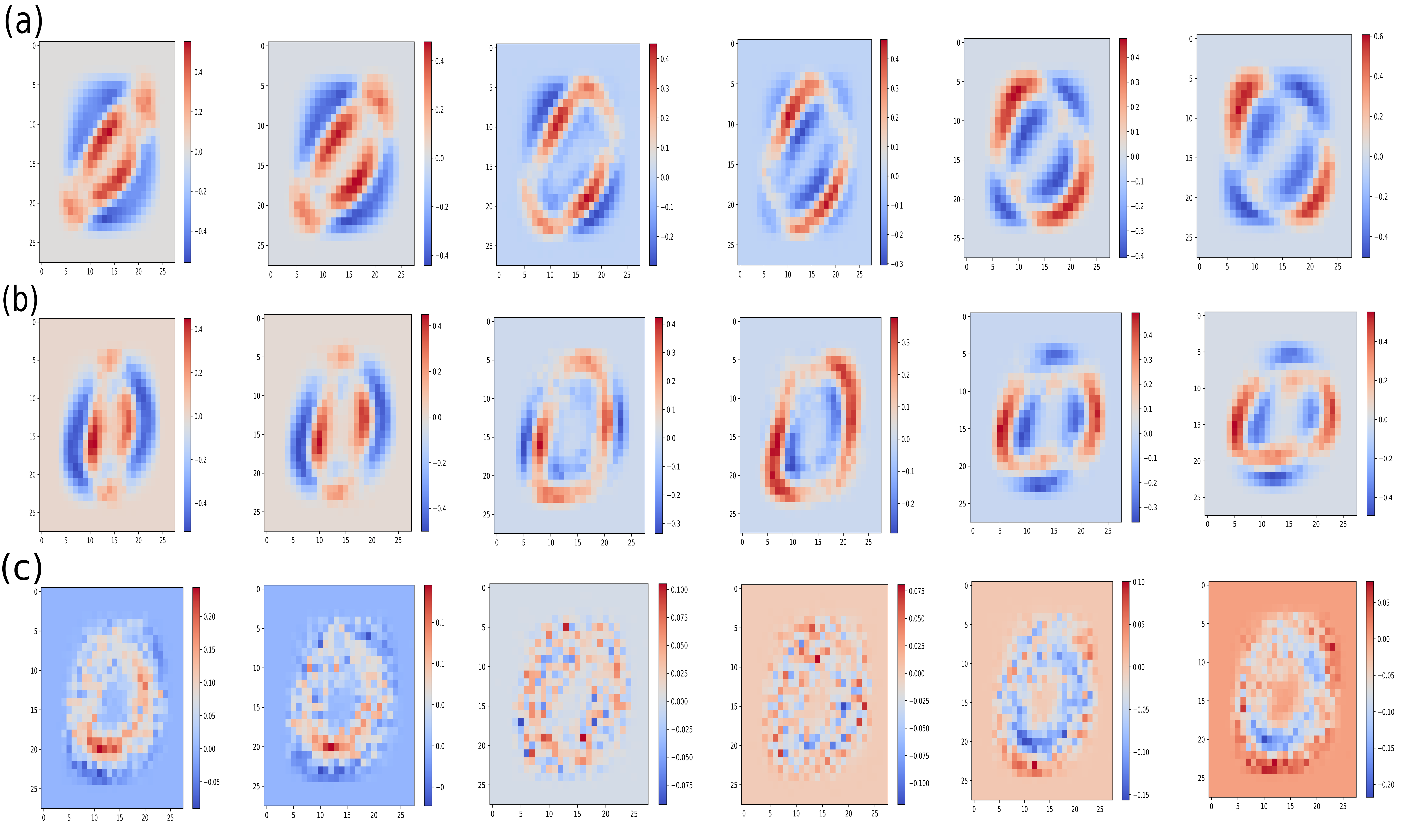}
  \caption{Causal attributions of (a) $z_0$ {\&} $c_0$, (b) $z_6$ {\&} $c_0$, (c) $z_2$ {\&} $c_0$ for the decoded image. Refer Section \ref{section: disentangled_latent_representations} for details. Red indicates a stronger causal effect, and blue indicates a weaker effect. The class-specific latent intervened on here is digit $0$.}
  \label{0_results}
\end{figure*}

\begin{figure*}
\centering
\includegraphics[scale=0.9]
{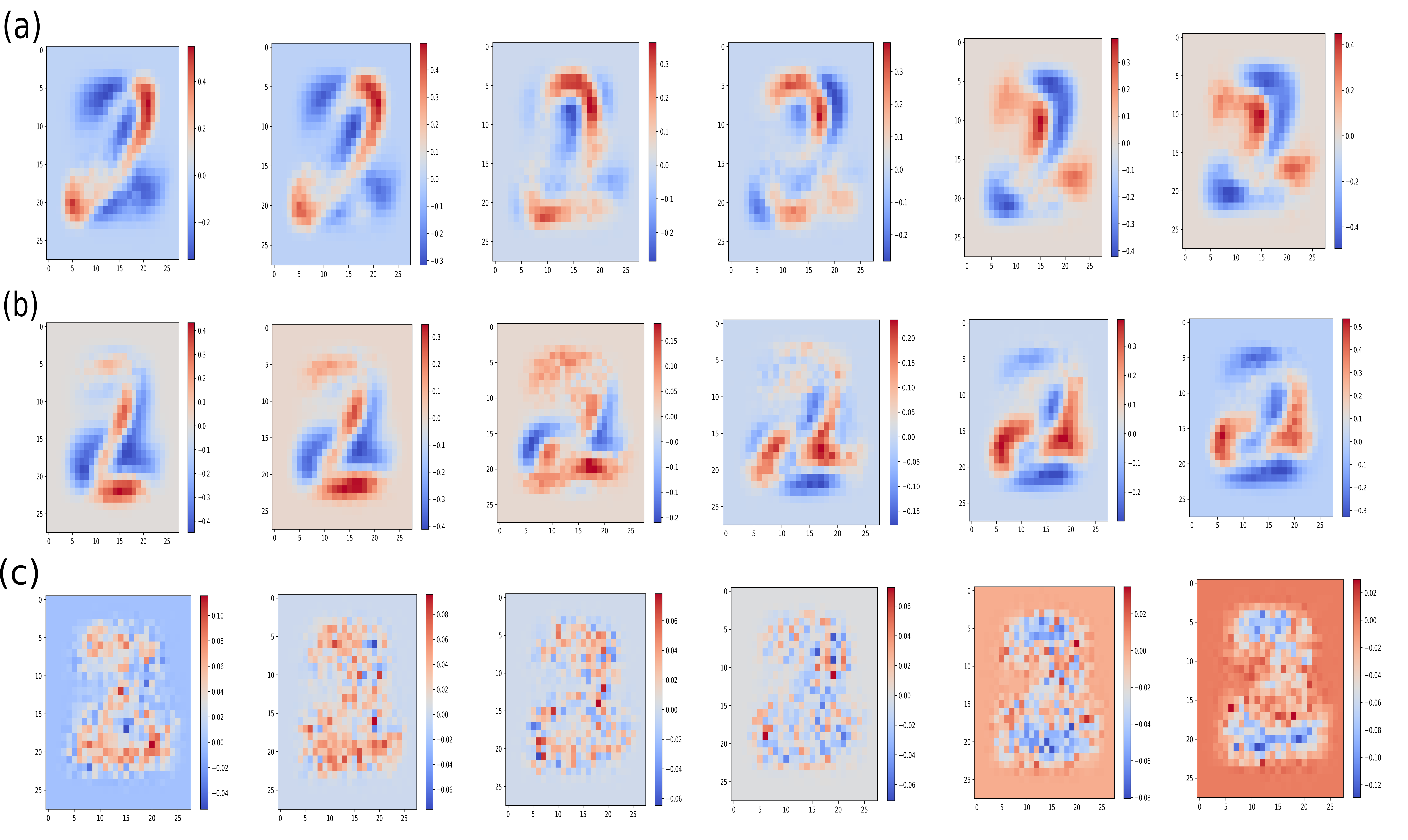}
  \caption{Causal attributions of (a) $z_0$ {$\&$} $c_3$, (b) $z_6$ {$\&$} $c_3$, (c) $z_2$ {$\&$} $c_3$ for the decoded image. Refer Section \ref{section: disentangled_latent_representations} for details. Red indicates a stronger causal effect, and blue indicates a weaker effect. The class-specific latent intervened on here is $2$.}
  \label{2_results}
\end{figure*}

\begin{figure*}
\centering
\includegraphics[scale=0.75]
{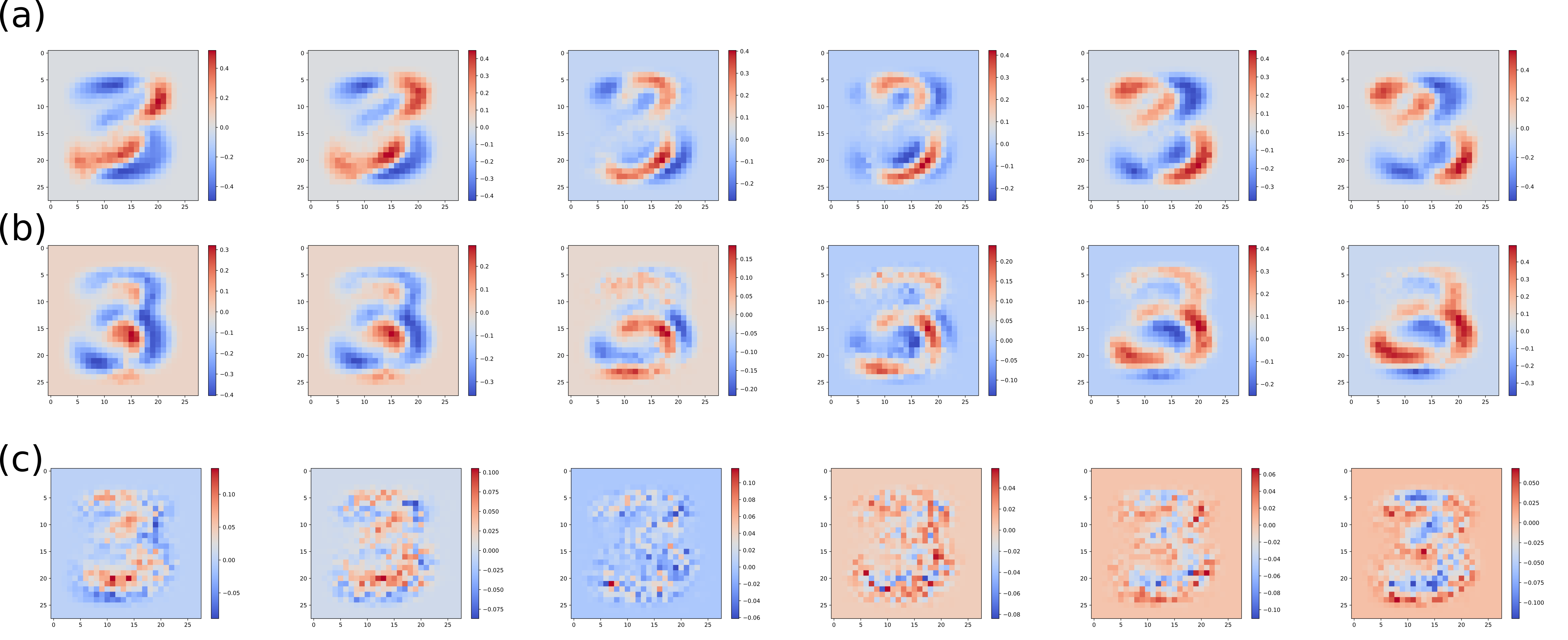}
  \caption{Causal attributions of (a) $z_0$ {$\&$} $c_3$, (b) $z_6$ {$\&$} $c_3$, (c) $z_2$ {$\&$} $c_3$ for the decoded image. Refer Section \ref{section: disentangled_latent_representations} for details. Red indicates a stronger causal effect, and blue indicates a weaker effect. The class-specific latent intervened on here is $3$.}
  \label{3_results}
\end{figure*}

Figure \ref{0_results} shows causal attributions of the continuous latents $z_k$ (defined in Section \ref{section: disentangled_latent_representations}) for the decoded image for different class-specific latents ($c_k$). In all the cases $z_0$ and $z_6$ capture rotation and scaling of the digit respectively. $z_2$ like all the other $z_k$s showed no discernable causal effect. We also show causal attributions of these latents for digits 0, 2 and 3 in Figure \ref{0_results} to \ref{3_results}.

\end{document}